\documentclass{article} 
\usepackage{emho_preprint,times}

\usepackage{amsmath,amsfonts,bm}

\def\eqref#1{equation~\ref{#1}}

\def\1{\bm{1}}

\DeclareMathAlphabet{\mathsfit}{\encodingdefault}{\sfdefault}{m}{sl}
\SetMathAlphabet{\mathsfit}{bold}{\encodingdefault}{\sfdefault}{bx}{n}

\usepackage[utf8]{inputenc}
\usepackage[T1]{fontenc}

\usepackage{graphicx}
\usepackage{float}
\usepackage{booktabs}
\usepackage{multirow}
\usepackage{array}
\usepackage{tabularx}
\usepackage{makecell}

\usepackage{amsfonts}
\usepackage{nicefrac}
\usepackage{microtype}
\usepackage{enumitem}

\usepackage[table]{xcolor}
\usepackage{subcaption}

\usepackage{url}
\usepackage{hyperref}

\makeatletter
\renewcommand{\@maketitle}{%
  \vbox{\hsize\textwidth\centering
    {\LARGE\scshape\@title\par}
    \begin{tabular}[t]{@{}c@{}}
      \rule{\z@}{24pt}\@author
    \end{tabular}
    \vskip 0.3in minus 0.1in
  }%
}
\makeatother

\title{EMHO: EMbodied Agent Harness Optimization via Experience Traces}

\author{%
\textbf{Hyun Jung Lee}\textsuperscript{1},
\textbf{Jungtaek Kim}\textsuperscript{2},
\textbf{Jongwon Jeong}\textsuperscript{3}, \\[0.25em]
\textbf{Tae-Eui Kam}\textsuperscript{1},
\textbf{Donghyun Kim}\textsuperscript{1,}\thanks{Corresponding authors.},
\textbf{Yong Jae Lee}\textsuperscript{3,*} \\[0.7em]
\normalfont\small\textsuperscript{1} Korea University \\
\normalfont\small\textsuperscript{2} University of Arkansas \\
\normalfont\small\textsuperscript{3} University of Wisconsin--Madison
}

\begin{document}

\maketitle

\begin{abstract}
Improving embodied agents often focuses on optimizing the underlying model through training, while the surrounding agent harness that controls planning, context, and tool use is typically engineered. We ask whether this harness can instead improve itself directly from experience traces under sparse environmental feedback. We propose \textbf{EMbodied Agent Harness Optimization (EMHO)}, a self-evolving framework that keeps the embodied model frozen and iteratively revises its harness by analyzing execution trajectories and prior harness history. EMHO optimizes beyond skills or recovery prompts, modifying how the agent monitors progress, uses vision tools, grounds observations, and responds to failures. To support multiple subtasks with a single harness, we introduce \textbf{EMHO-Merge}, which addresses trade-offs in jointly optimizing a single shared harness across subtasks by using episode-level gains and losses to guide evidence-supported refinement of when and how revised behaviors are applied. We evaluate EMHO on EmbodiedBench across navigation and manipulation tasks, and EMHO consistently improves task success for both Qwen 9B and 27B models. Qualitative analysis shows that EMHO goes beyond recovering from failures and unproductive actions to reshape how the embodied agent interprets and interacts with its environment.
The code can be found \href{https://hyunjung00.github.io/emho/}{here}.
\end{abstract}

\section{Introduction}

Recent progress in embodied agents has largely been driven by advances in the underlying vision-language models~\citep{palme, pi05, okrobot, openvla}. Yet a better embodied agent does not always require a better model. Even when the underlying vision-language model is frozen, changes to the surrounding agent system can substantially alter what the agent observes, how it reasons, which actions it attempts, and how it responds when those actions fail. This surrounding system is referred to as \emph{harness}: the logic that organizes planning, context and memory, tool selection, observation presentation, and failure recovery. 

Designing a strong embodied-agent harness, however, remains largely manual because failures are difficult to diagnose from interaction traces. An episode may span many observations, plans, actions, and tool calls, while the final outcome provides little indication of which earlier decision caused the failure. This contrasts with coding and command-line environments, where failed tests, exceptions, or localized execution errors can directly guide self-improvement~\citep{selfharness, metaharness}. Embodied agents instead often receive only sparse action-level validity feedback and a delayed episode-level success or failure signal, making it more difficult and costly to identify recurring weaknesses and determine how the harness should be revised.

Recent work on embodied self-improvement has focused on refining specific components of the agent, such as reusable skills or recovery mechanisms~\citep{ju2026embodiskill,yang2026vaso,lu2026aspire}. While these approaches can improve particular behaviors or responses to failure, an embodied agent's behavior is determined by a broader set of interacting mechanisms, including planning, context and memory construction, external vision tool usage, visual grounding and failure recovery. When a weakness arises from one of these surrounding mechanisms, updating only a skill or recovery component may leave the underlying source of the problem unchanged. This raises a question: \emph{can an embodied agent improve this broader harness itself from sparse and weakly diagnostic interaction feedback?} We introduce \textbf{EMbodied Agent Harness Optimization} (\textbf{EMHO}), a self-evolving framework that keeps the embodied model frozen and iteratively optimizes the surrounding harness. In the inner loop, the agent executes embodied tasks and records trajectories containing observations, plans, actions, tool usage, environmental feedback, and final outcomes. In the outer loop, a harness optimizer analyzes these trajectories together with the history of previous harness modifications and their associated outcomes, identifies recurring behavioral weaknesses, and proposes targeted edits to the harness. 


To further explore whether one harness can perform well across related subtasks, we extend EMHO from subtask-specific optimization to joint optimization over a combined search set. When optimizing a single harness across multiple subtasks, a revision that improves performance on one subtask may reduce performance on another. We introduce \textbf{EMHO-Merge} to use these gains and losses to guide further refinement. It compares each candidate with its parent harness to identify episodes that change from failure to success and those that change from success to failure. By examining the corresponding trajectories before and after the revision, it looks for observable conditions that distinguish when the revised behavior helps from when it hurts. When supported by this evidence, the optimizer adjusts when or how the behavior is applied, aiming to retain the gains while reducing the losses.

We evaluate EMHO on EmbodiedBench~\citep{embodiedbench} across navigation and manipulation tasks. EMHO substantially improves task success over the initial harness, with gains of up to 23.3\% in navigation and 33.3\% in manipulation. When optimizing a single shared harness across subtasks, EMHO-Merge achieves higher average search set success than EMHO on the pooled subtasks in both navigation and manipulation. Beyond improving failure recovery, our analysis shows that EMHO reshapes how the agent perceives and acts in the environment, changing how it monitors progress, orchestrates perceptual tools, and grounds visual observations into subsequent actions.

\begin{figure}[t]
    \centering
    \includegraphics[width=\linewidth]{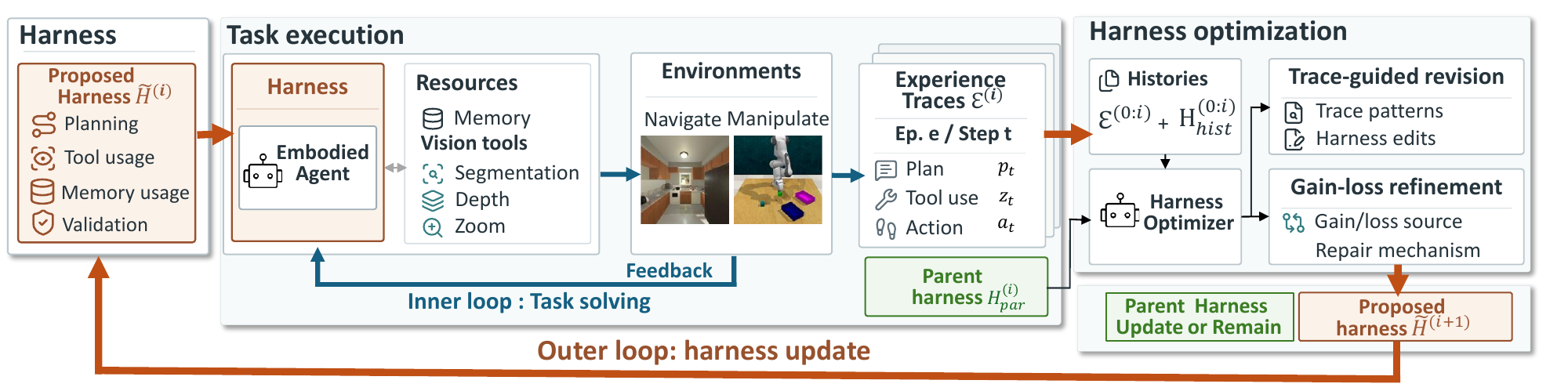}
    \caption{Overview of EMHO and EMHO-Merge. During harness optimization, EMHO uses trace-guided revision, while EMHO-Merge adds gain--loss refinement.} 
    
    \label{fig:overview}
\end{figure}

Our contributions are summarized as follows:
\begin{itemize}[leftmargin=*]

\item \textbf{Self-evolving harness for embodied agents via EMHO.}
We suggest \textsc{EMHO}, which evolves the broader harness around a frozen model from interaction traces under sparse and weakly diagnostic environmental feedback, moving beyond adaptation restricted to skills or recovery instructions. 

\item \textbf{Shared harness optimization via EMHO-Merge.}
We extend EMHO to optimize a single shared harness across subtasks toward more general-purpose embodied behavior. EMHO-Merge explicitly contrasts episode-level gains and losses and uses paired trajectories to guide evidence-supported repairs, aiming to retain useful adaptations while mitigating trade-offs during joint optimization.

\item \textbf{EMHO's ability to reshape agents' behavior.}
EMHO improves task success, and qualitative analysis shows that the evolved harness improves failure recovery and reshapes how the agent monitors progress, uses perceptual tools, and grounds visual observation.

\end{itemize}

\section{Related Work}

\paragraph{Coding-Agent Harness Optimization.}
Recent work has begun to automate the optimization of coding-agent harnesses using execution traces and evaluation outcomes to revise prompts and skills. Meta-Harness~\citep{metaharness} searches over executable harness code using the source code, execution traces, and scores of previously evaluated candidates, while Self-Harness~\citep{selfharness} identifies model-specific weaknesses and proposes targeted modifications that are retained through regression-based validation. Related approaches improve observability over editable agent components~\citep{agenticharnessengineer} or combine failure diagnosis and validation to produce persistent harness updates~\citep{autosaddler}. These methods have primarily been studied in coding and command-line environments, where failures often provide relatively explicit diagnostic signals such as failed tests, exceptions, or localized execution errors. Embodied environments, in contrast, may provide only sparse action-level feedback during interaction and a delayed task-level success signal.

\paragraph{Self-Improvement in Embodied Agents.}
Embodied-agent research has similarly explored improving behavior from interaction experience while keeping the underlying model fixed. Existing approaches concentrate self-improvement on a specific behavioral component, such as reusable skills, task-specific policies, or recovery mechanisms. EmbodiSkill~\citep{ju2026embodiskill} revises skills through trajectory-based reflection, VASO~\citep{yang2026vaso} uses verification and counterexamples to guide skill evolution, ASPIRE~\citep{lu2026aspire} improves robotic behavior by iteratively evolving task-specific code and accumulating successful repairs as reusable skills, and Zetta~\citep{ding2026zetta} focuses on failure detection and recovery by evolving runtime critics together with corresponding recovery behaviors. Our work extends this direction by treating the broader harness as the object of optimization, rather than focusing on a particular skill or recovery component. We further study how a single shared harness can be optimized across subtasks with differing behavioral demands.

\section{Embodied Agent Harness Optimization (EMHO)}
\label{sec:method}

We optimize the harness $H$ surrounding a frozen vision-language model $M$. The harness consists of the agent-side instructions and runtime logic that structure interaction with the environment, including observation and context construction, memory management, planning, tool use, action generation, output validation, and failure recovery. Let $\mathcal{H}$ denote the space of possible harnesses. Our objective is to find a harness that maximizes expected task success:
\begin{equation}
H^\star
=
\operatorname*{arg\,max}_{H \in \mathcal{H}}
\mathbb{E}_{e \sim \mathcal{D}}
\left[
    y_e(M,H)
\right],
\label{eq:objective}
\end{equation}
where $y_e(M,H) \in \{0,1\}$ denotes the final outcome of episode $e$. Throughout optimization, only the harness is modified while the embodied model and environment remain fixed. EMHO alternates between two stages: an inner loop that executes the current harness on embodied tasks and records experience traces, and an outer loop that analyzes these traces alongside the history of previous optimization attempts to propose a targeted harness revision.

\subsection{EMHO}
\paragraph{Inner Loop: Experience Trace Collection}
\label{sec:inner_loop}

At optimization iteration $i$, the agent executes the current harness $H^{(i)}$ on each episode $e \in \mathcal{D}_{\mathrm{search}}$. The resulting execution trace is recorded as
\begin{equation}
\mathcal{T}_e^{(i)} = \left( x_e,\, \left\{ (o_t,p_t,z_t,a_t,f_t) \right\}_{t=1}^{T_e},\, y_e \right),
\quad
\mathcal{E}^{(i)} = \left\{ \mathcal{T}_e^{(i)} \mid e \in \mathcal{D}_{\mathrm{search}} \right\}.
\end{equation}
Here, $x_e$ specifies the task and $y_e$ denotes the final episode outcome. At interaction step $t$, $o_t$, $p_t$, $z_t$, $a_t$, and $f_t$ denote the observation, generated plan, optional tool interaction, executed action, and environmental feedback, respectively. Embodied feedback is often sparse and weakly diagnostic, as a failed action or unsuccessful episode does not directly reveal the underlying cause.

\paragraph{Outer Loop: History-Aware Harness Refinement}
\label{sec:outer_loop}

The outer loop maintains a current parent harness $H_{\mathrm{par}}^{(i)}$, which serves as the basis for the next revision. The optimizer also receives the complete experience and harness histories accumulated through iteration $i$. It proposes a candidate harness and updates the parent according to
\begin{equation}
\widetilde{H}^{(i+1)}
=
\mathcal{O}
\left(
    H_{\mathrm{par}}^{(i)},
    \mathcal{E}^{(0:i)},
    \mathbf{H}_{\mathrm{hist}}^{(0:i)}
\right),
\quad
H_{\mathrm{par}}^{(i+1)}
=
\begin{cases}
\widetilde{H}^{(i+1)},
&
\text{if} \
s\left(\widetilde{H}^{(i+1)}\right)
>
s\left(H_{\mathrm{par}}^{(i)}\right),
\\
H_{\mathrm{par}}^{(i)},
&
\text{otherwise}.
\end{cases}
\label{eq:harness_optimization}
\end{equation}
Here, $s(H)
=
\frac{1}{|\mathcal{D}_{\mathrm{search}}|}
\sum_{e \in \mathcal{D}_{\mathrm{search}}}
y_e(M,H)$
denotes the success rate of harness $H$ on the search set.
$\mathcal{E}^{(0:i)}$ denotes all experience traces collected from the initial evaluation through iteration $i$, including traces from both accepted and rejected candidates. Similarly, $\mathbf{H}_{\mathrm{hist}}^{(0:i)}$ denotes the complete harness history, including the previously evaluated harnesses, the modifications applied to them, and their evaluation outcomes. Using these histories, the optimizer identifies recurring behavioral weaknesses and proposes a targeted modification of the current parent harness. The candidate is evaluated and replaces the parent only if it achieves a higher search-set success rate. An example of the harness optimizer prompt is provided in Appendix~\ref{appendix:harness_structure}.

\subsection{EMHO-Merge}
\label{sec:EMHO_Unify}

\paragraph{Gain--Loss-Aware Refinement.}
To move toward a more general-purpose embodied agent, we optimize a single shared harness over the combined search set of multiple subtasks. Since a revision can improve some episodes while degrading others, EMHO-Merge uses these mixed outcomes to guide refinement of when and how the revised behavior is applied. EMHO-Merge extends the outer loop of EMHO by using episode-level outcome changes from the most recently evaluated candidate to guide the next proposal. The candidate $\widetilde{H}^{(i)}$ is compared with its parent harness $H_{\mathrm{par}}^{(i-1)}$. Episodes that change from failure to success are gains $\mathcal{G}^{(i)}$, while those that change from success to failure are losses $\mathcal{L}^{(i)}$. When both occur, the optimizer may repair the previous harness revision by adjusting how or when the behavior is applied, rather than introducing a different behavior. It examines paired gain and loss traces to identify observable conditions under which the behavior is expected to improve task success, aiming to preserve the gains while reducing the losses. Each original revision is allowed at most one evaluated repair, which is proposed and evaluated as the candidate in the next optimization iteration. The optimizer determines whether repair is available and proposes the next candidate as
\begin{equation}
r^{(i)}
\!=\!
\mathbb{I}\!\left[
|\mathcal{G}^{(i)}|\!>\!0
\land
|\mathcal{L}^{(i)}|\!>\!0
\land
\text{repair unused}
\right],
\ \
\widetilde{H}^{(i+1)}
\!=\!
\mathcal{O}\!\left(
H_{\mathrm{par}}^{(i)},
\mathcal{E}^{(0:i)},
\mathcal{G}^{(i)},
\mathcal{L}^{(i)};
r^{(i)}
\right).
\label{eq:emho_unify}
\end{equation}
Here, $r^{(i)}=1$ indicates that the revision evaluated in candidate $\widetilde{H}^{(i)}$ is available for repair in the next iteration. If repair is unavailable or unsupported by the collected experience, the optimizer proposes a different behavioral change. The next candidate $\widetilde{H}^{(i+1)}$ is constructed from the current parent $H_{\mathrm{par}}^{(i)}$, allowing experience from a rejected candidate to guide the next revision.

\begin{table}[t]
\centering
\caption{
Success rates on the navigation and manipulation tasks for each subtask. Initial harness and EMHO results are reported separately for the search and held-out splits. EMHO entries report the best score attained on each split over the optimization iterations. Bold entries denote the highest score.  More details on the baseline and subtask-level results are provided in Appendix~\ref{appendix:quantitative_result}.}
\label{tab:main}

\scriptsize
\setlength{\tabcolsep}{10pt}
\renewcommand{\arraystretch}{1.05}



\begin{tabular}{lcccccc}
\toprule
\multirow{2}{*}{\textbf{Method / Model}} &
\multicolumn{6}{c}{\textbf{Navigation}} \\
\cmidrule(lr){2-7}
&
\textbf{Avg} &
\textbf{Base} &
\textbf{Common} &
\textbf{Complex} &
\textbf{Visual} &
\textbf{Long} \\
\midrule

GPT-5
& 56.8
& 62.0
& 62.0
& 68.0
& 50.0
& 42.0 \\


Claude-4.6-Sonnet
& 62.2
& 68.0
& 63.0
& \textbf{77.0}
& 55.0
& 48.0 \\


Gemini-2.0-flash
& 48.7
& 63.3
& 65.0
& 50.0
& 51.7
& 13.3 \\

InternVL3-78B
& 53.7
& 66.7
& 63.3
& 61.7
& 45.0
& 31.7 \\

gemma-3-27b-it
& 45.3
& 53.3
& 45.0
& 61.7
& 50.0
& 16.7 \\

\midrule

One-Shot Harness w/ Opus (Qwen3.5-9B)
& 53.3
& 56.7
& 60.0
& 63.3
& 36.7
& 50.0 \\

One-Shot Harness w/ Sol (Qwen3.5-9B)
& 49.3
& 50.0
& 53.3
& 63.3
& 40.0
& 40.0 \\

One-Shot Harness w/ Opus (Qwen3.8-27B)
& 67.7
& 78.3
& 70.0
& 75.0
& 58.3
& 56.7 \\

One-Shot Harness w/ Sol (Qwen3.8-27B)
& 65.7
& 73.3
& 70.0
& 61.7
& 63.3
& 60.0 \\

\midrule

Initial Harness (Search, Qwen3.5-9B)
& 50.0
& 58.3
& 46.7
& 65.0
& 40.0
& 40.0 \\

Initial Harness (Held-out, Qwen3.5-9B)
& 48.7
& 58.3
& 53.3
& 58.3
& 33.3
& 40.0 \\

EMHO (Search, Qwen3.5-9B)
& 63.0
& 71.7
& 70.0
& 70.0
& 50.0
& 53.3 \\

EMHO (Held-out, Qwen3.5-9B)
& 56.7
& 66.7
& 60.0
& 65.0
& 40.0
& 51.7 \\

\midrule

Initial Harness (Search, Qwen3.8-27B)
& 54.7
& 60.0
& 60.0
& 50.0
& 50.0
& 53.3 \\

Initial Harness (Held-out, Qwen3.8-27B)
& 66.0
& 73.3
& 63.3
& 66.7
& 66.7
& 60.0 \\

EMHO (Search, Qwen3.8-27B)
& 67.3
& 76.7
& \textbf{76.7}
& 63.3
& 63.3
& 56.7 \\

EMHO (Held-out, Qwen3.8-27B)
& \textbf{75.3}
& \textbf{86.7}
& \textbf{76.7}
& 73.3
& \textbf{76.7}
& \textbf{63.3} \\

\bottomrule
\end{tabular}

\vspace{3pt}


\begin{tabular}{lcccccc}
\toprule
\multirow{2}{*}{\textbf{Method / Model}} &
\multicolumn{6}{c}{\textbf{Manipulation}} \\
\cmidrule(lr){2-7}
&
\textbf{Avg} &
\textbf{Base} &
\textbf{Common} &
\textbf{Complex} &
\textbf{Visual} &
\textbf{Spatial} \\
\midrule

GPT-5
& 26.8
& 21.0
& 25.0
& 31.0
& 22.0
& 35.0 \\


Claude-4.6-Sonnet
& 37.6
& 40.0
& 35.0
& \textbf{40.0}
& 36.0
& 37.0 \\


Gemini-2.0-flash
& 16.5
& 14.6
& 8.3
& 14.6
& 13.9
& 31.3 \\

InternVL3-78B
& 26.3
& 29.2
& 22.9
& 22.9
& 25.0
& 31.3 \\

gemma-3-27b-it
& 17.5
& 25.0
& 16.7
& 16.7
& 8.3
& 20.8 \\

\midrule

One-Shot Harness w/ Opus (Qwen3.5-9B)
& 13.2
& 10.4
& 16.7
& 12.5
& 13.9
& 12.5 \\

One-Shot Harness w/ Sol (Qwen3.5-9B)
& 11.8
& 12.5
& 6.3
& 10.4
& 11.1
& 18.8 \\

One-Shot Harness w/ Opus (Qwen3.8-27B)
& 27.8
& 22.9
& 20.8
& 29.2
& 38.9
& 27.1 \\

One-Shot Harness w/ Sol (Qwen3.8-27B)
& 25.3
& 25.0
& 25.0
& 27.1
& 22.2
& 27.1 \\

\midrule

Initial Harness (Search, Qwen3.5-9B)
& 9.7
& 8.3
& 8.3
& 6.3
& 11.1
& 14.6 \\

Initial Harness (Held-out, Qwen3.5-9B)
& 10.7
& 6.3
& 4.2
& 8.3
& 22.2
& 12.5 \\

EMHO (Search, Qwen3.5-9B)
& 18.3
& 14.6
& 12.5
& 16.7
& 25.0
& 22.9 \\

EMHO (Held-out, Qwen3.5-9B)
& 21.0
& 14.6
& 20.8
& 18.8
& 36.1
& 14.6 \\

\midrule

Initial Harness (Search, Qwen3.8-27B)
& 23.9
& 16.7
& 12.5
& 29.2
& 44.4
& 16.7 \\

Initial Harness (Held-out, Qwen3.8-27B)
& 26.4
& 25.0
& 25.0
& 29.2
& 27.8
& 25.0 \\

EMHO (Search, Qwen3.8-27B)
& \textbf{44.2}
& \textbf{50.0}
& \textbf{41.7}
& 37.5
& \textbf{50.0}
& \textbf{41.7} \\

EMHO (Held-out, Qwen3.8-27B)
& 39.4
& 41.7
& \textbf{41.7}
& 33.3
& 38.9
& \textbf{41.7} \\

\bottomrule
\end{tabular}
\end{table}

\section{Quantitative Results and Discussion}

We conduct experiments on EmbodiedBench~\citep{embodiedbench}, covering both navigation and manipulation tasks. We use the Qwen3.5-9B model~\citep{qwen3.5} and the Qwen3.8-27B-FP8 model~\citep{qwen3.8} as the underlying embodied agents. We evaluate both models in the main results and use the 27B model for all other experiments unless otherwise stated. For visual perception, the agent is equipped with Depth Anything 3~\citep{depthv3} for depth estimation, SAM 3~\citep{SAM3} for segmentation, and zoom tools. We initialize the agent harness based on the original EmbodiedBench implementation and perform 10 iterations of harness optimization for each task. 
For each task, we divide the benchmark evenly into disjoint search and held-out sets. We use the search set to collect experience traces and optimize the harness. The harnesses produced during optimization are also evaluated on the held-out set. Details of the harness structure are provided in Appendix~\ref{appendix:harness_structure}.

\subsection{Main Results}

As shown in Table~\ref{tab:main}, EMHO improves performance on the search set for navigation and manipulation, and the evolved harnesses also achieve higher success than the initial harness on held-out episodes. The improvements also extend beyond the relatively simple Base tasks, appearing across diverse subtasks that require longer-horizon reasoning, visual grounding, and state-aware manipulation. Notably, the gains are observed for both the 9B and 27B models, the evolved 9B navigation harness reaches performance competitive with larger published models on search set, while the 27B model also benefits considerably from harness evolution, particularly in manipulation. To further analyze these gains, we ablate the contributions of experience traces and harness-revision history. As reported in Appendix~\ref{appendix:history_ablation}, both sources of context improve EMHO's performance.

\subsection{Comparison with Frontier-Model One-Shot Revision}

We use one-shot harness revisions generated by Claude Opus-5~\citep{opus5} and GPT-5.6 Sol~\citep{gpt56sol} as baselines to assess the gains achieved by EMHO.
For each domain, these models receive the initial harness provided in EmbodiedBench,  representative task examples, and are asked to produce a single improved harness. Although one-shot revision improves the initial harness, EMHO performs better across both domains. One-shot revisions tend to provide broad guidance, such as changing strategy when a loop occurs, without specifying how the agent should recognize and respond to the failure. EMHO instead converts recurring patterns into concrete operating conditions. For example, when a movement is blocked, the evolved harness terminates the remaining actions in the current plan, avoids retrying the same direction, and replans from the updated observation. This example illustrates how interaction experience helps EMHO develop specific execution rules beyond the broad guidance of one-shot revisions.

\subsection{Full-Harness versus Skill-Only Optimization}
Skill-only optimization primarily improves behavior by accumulating and refining textual recovery guidance. Across iterations, the learned skills become increasingly specific about how to respond to recurring failures, such as blocked navigation or repeated ineffective actions. Full-harness optimization discovers similar recovery principles, but is not limited to expressing them as instructions: it can also modify how the interaction history is represented, add mechanisms that explicitly detect repeated or failed behavior, and ensure that the resulting state is exposed to subsequent planning. Moreover, it can intervene beyond recovery by changing progress monitoring, tool orchestration, and grounding behavior. This broader optimization space is reflected in the substantially larger average performance gains of full-harness optimization in Table~\ref{tab:skill_only}. The comparison suggests that the benefit of harness evolution comes not only from discovering better behavioral rules, but also from changing the surrounding system so that those rules can be detected, triggered, and enforced during interaction.

\begin{figure}[t]
    \centering
    \includegraphics[width=\linewidth]{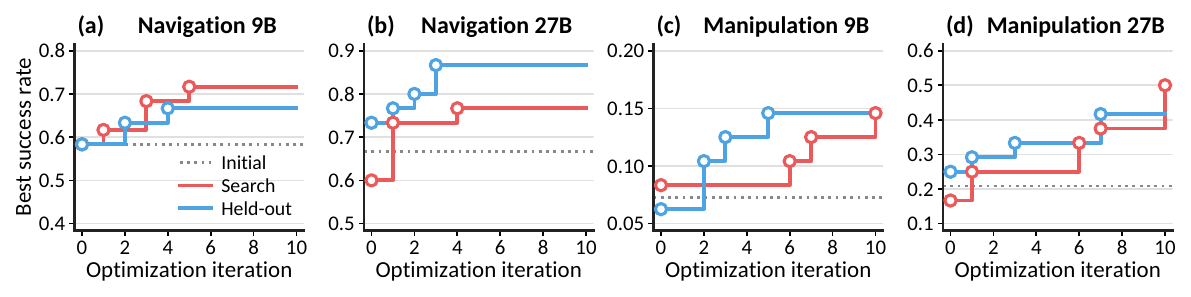}
    \caption{Quantitative results on the navigation and manipulation base subtask across optimization iterations. Full results are reported in Appendix~\ref{appendix:quantitative_result}.}
    \label{fig:base_result}
\end{figure}

\subsection{Shared Harness Optimization with EMHO-Merge}

When optimizing a shared harness across subtasks, a revision that helps one subtask may interfere with another. Separately optimized navigation harnesses often evolve different responses to blocked movement: Base favors an immediate action change, visual-appearance prefers directions supported by visible open space, and long-horizon first reassesses whether a direction change is necessary. Simply pooling experience across these subtasks can therefore introduce conflicting behaviors. For example, a rule that moves laterally before proceeding forward after a blocked or repeated approach helped Base navigation, but repeatedly triggered in a long-horizon trajectory, redirecting the agent into nearby obstacles.
We evaluate EMHO-Merge on three subtasks per domain against EMHO (Separate) and EMHO (Combined), as shown in Table~\ref{tab:merged_harness}. EMHO-Merge uses gain--loss trajectories to refine when a behavior should activate; for instance, obstacle avoidance is restricted to cases with evidence of blockage. It achieves higher average success than EMHO (Combined) in both navigation and manipulation, suggesting that gain--loss-aware refinement better manages cross-subtask trade-offs.

\begin{table}[t]
\centering
\caption{Success-rate gains in percentage points over the initial harness on the search set, comparing skill-only updates with full-harness optimization using EMHO.}
\label{tab:skill_only}


\setlength{\tabcolsep}{2pt}
\renewcommand{\arraystretch}{1.05}

\begin{subtable}[t]{0.49\linewidth}
\centering
\caption{Navigation}
\scriptsize
\begin{tabularx}{\linewidth}{@{}>{\raggedright\arraybackslash}p{0.16\linewidth}*{5}{>{\centering\arraybackslash}X}@{}}
\toprule
\textbf{Method}
& \textbf{Base}
& \textbf{Common}
& \textbf{Complex}
& \textbf{Visual}
& \textbf{Long} \\
\midrule
Skill-only
& +10.0
& +10.0
& 0.0
& +10.0
& +3.3 \\
EMHO
& +16.7
& +16.7
& +13.3
& +13.3
& +3.3 \\
\bottomrule
\end{tabularx}
\end{subtable}
\hfill
\begin{subtable}[t]{0.49\linewidth}
\centering
\caption{Manipulation}
\scriptsize
\begin{tabularx}{\linewidth}{@{}>{\raggedright\arraybackslash}p{0.16\linewidth}*{5}{>{\centering\arraybackslash}X}@{}}
\toprule
\textbf{Method}
& \textbf{Base}
& \textbf{Common}
& \textbf{Complex}
& \textbf{Visual}
& \textbf{Spatial} \\
\midrule
Skill-only
& +16.7
& +8.3
& +8.3
& +11.1
& +8.3 \\
EMHO
& +33.3
& +29.2
& +8.3
& +5.6
& +25.0 \\
\bottomrule
\end{tabularx}
\end{subtable}

\end{table}

\begin{table}[t]
\centering
\caption{
Subtask-wise comparison of different harness optimization settings on the
search and held-out sets. EMHO (Separate) refers to independent optimization
for each subtask, whereas EMHO (Combined) optimizes a single shared harness
using experience traces pooled across subtasks.
}
\label{tab:merged_harness}

\begin{subtable}[t]{0.49\linewidth}
\vspace{0pt}
\centering
\caption{Navigation}
\scriptsize

\makebox[\linewidth][c]{%
\scalebox{0.82}{%
\begin{tabular}{llcccc}
\toprule
\textbf{Split} & \textbf{Harness} & \textbf{Avg.}
& \textbf{Base} & \textbf{Visual} & \textbf{Long} \\
\midrule

\multirow{4}{*}{Search}
& Initial         & 54.4 & 60.0 & 50.0 & 53.3 \\
& EMHO (Separate) & 65.6 & 76.7 & 63.3 & 56.7 \\
& EMHO (Combined) & 61.1 & 73.3 & 63.3 & 46.7 \\
& EMHO-Merge      & 67.8 & 80.0 & 60.0 & 63.3 \\
\midrule

\multirow{3}{*}{Held-out}
& Initial         & 66.7 & 73.3 & 66.7 & 60.0 \\
& EMHO (Separate) & 75.6 & 86.7 & 76.7 & 63.3 \\
& EMHO-Merge      & 75.6 & 93.3 & 70.0 & 63.3 \\

\bottomrule
\end{tabular}%
}}
\end{subtable}
\hfill
\begin{subtable}[t]{0.49\linewidth}
\vspace{0pt}
\centering
\caption{Manipulation}
\scriptsize

\makebox[\linewidth][c]{%
\scalebox{0.82}{%
\begin{tabular}{llcccc}
\toprule
\textbf{Split} & \textbf{Harness} & \textbf{Avg.}
& \textbf{Base} & \textbf{Visual} & \textbf{Spatial} \\
\midrule

\multirow{4}{*}{Search}
& Initial         & 25.9 & 16.7 & 44.4 & 16.7 \\
& EMHO (Separate) & 47.2 & 50.0 & 50.0 & 41.7 \\
& EMHO (Combined) & 34.7 & 41.7 & 33.3 & 29.2 \\
& EMHO-Merge      & 39.8 & 37.5 & 44.4 & 37.5 \\
\midrule

\multirow{3}{*}{Held-out}
& Initial         & 25.9 & 25.0 & 27.8 & 25.0 \\
& EMHO (Separate) & 40.7 & 41.7 & 38.9 & 41.7 \\
& EMHO-Merge      & 33.8 & 29.2 & 38.9 & 33.3 \\

\bottomrule
\end{tabular}%
}}
\end{subtable}

\end{table}

\begin{figure}[t]
    \centering
    \includegraphics[width=\linewidth]{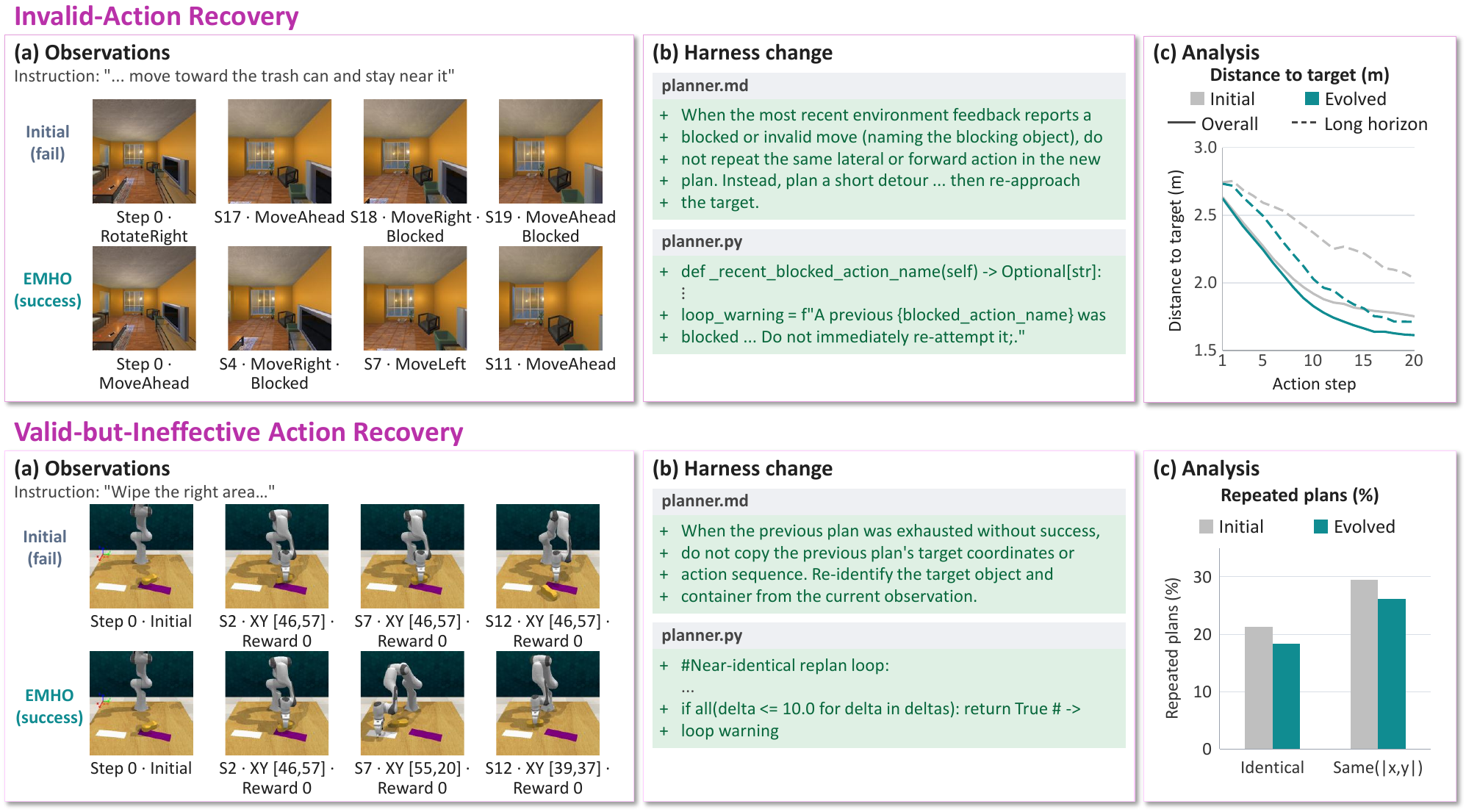}
    \caption{Qualitative analysis of recovery edits during harness optimization. (a) Representative observation frames from the initial harness and evolved EMHO trajectories. (b) Examples of the corresponding harness changes. (c) Analysis of their effects.} 
    
    \label{fig:qual_recovery}
\end{figure}

\begin{figure}[t]
    \centering
    \includegraphics[width=\linewidth]{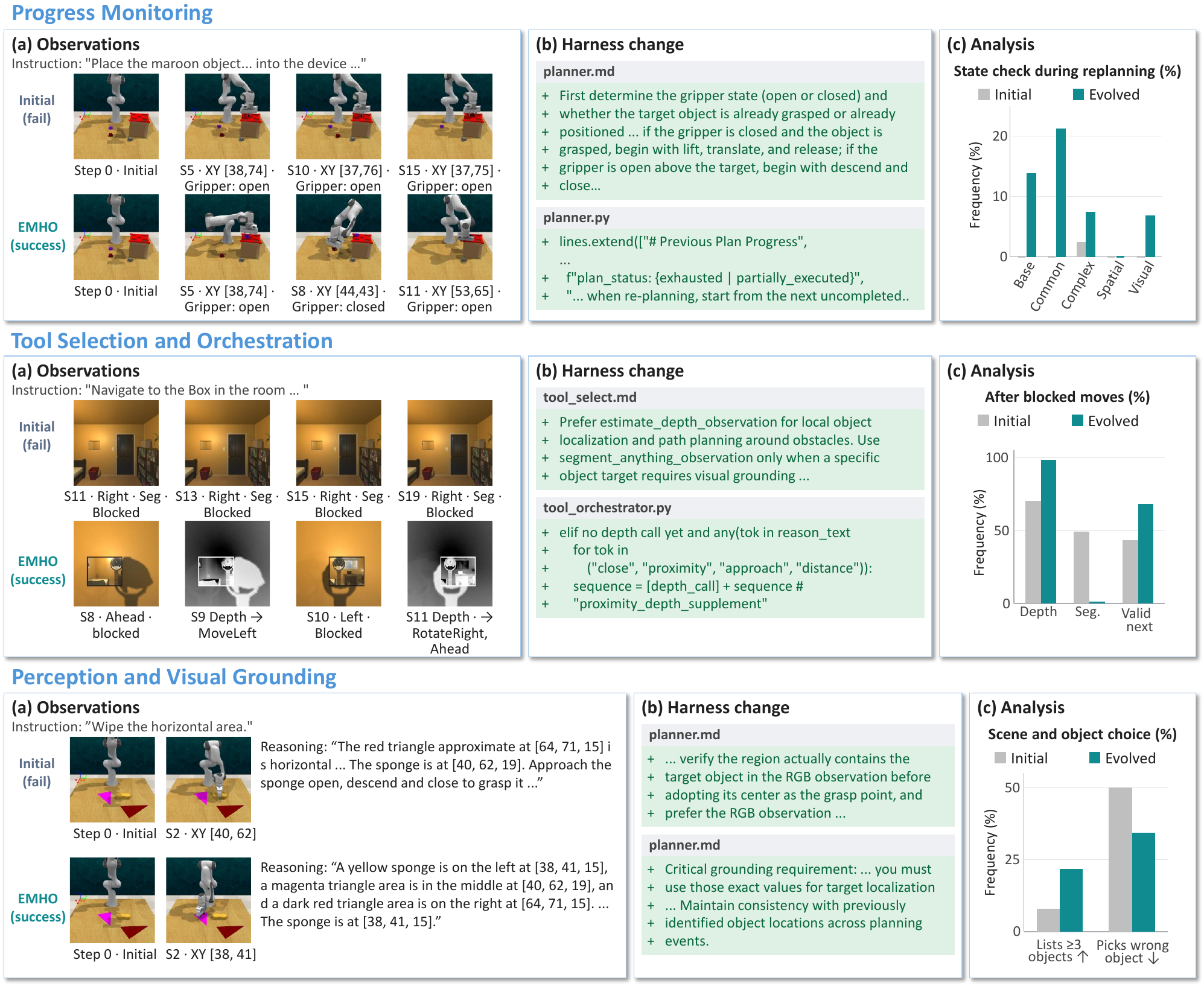}
    \caption{Qualitative analysis of non-recovery edits during harness optimization. (a) Representative observation frames from the initial harness and evolved EMHO trajectories. (b) Examples of the corresponding harness changes. (c) Analysis of their effects.} 
    \label{fig:qual_non_recovery}
\end{figure}

\section{Qualitative Analysis of Harness Evolution}
\label{sec:qualitative_results}

Aggregate success rates show whether harness optimization improves performance, but not how the resulting behavior changes. We therefore analyze the edits discovered by EMHO and their effects on held-out trajectories, focusing primarily on results from the 27B model. We categorize the resulting edits into \textbf{recovery} and
\textbf{non-recovery} edits. Recovery edits determine how the agent responds
when execution fails or ceases to make progress, whereas non-recovery edits
change how the agent monitors progress, selects external vision tools, and
grounds decisions during ongoing interaction. For each behavior, we examine the corresponding harness change together with its effect across held-out trajectories. Due to space constraints, we illustrate each modification type with an example from either navigation or manipulation. The full set of examples and analyses is provided in Appendix~\ref{appendix:qualitative_results}.

\subsection{Recovery Edits}
\label{sec:qual_recovery}

\noindent\textbf{Invalid-Action Recovery.}
Embodied environments provide sparse execution feedback, such as blocked navigation movements or unsuccessful manipulation actions. The initial harness provides limited guidance on how to respond to these failures, so the agent may repeat unsuccessful actions. EMHO adds explicit recovery instructions and modifies the harness to retain failure feedback and carry it into the context for subsequent planning, as illustrated in Figure~\ref{fig:qual_recovery}. In navigation, the evolved harness instructs the agent to avoid retrying a blocked movement and instead reorient or take a short detour. Consistent with these changes, evolved trajectories are more likely to continue reducing their distance to the target after encountering a blockage. Note that the distance to the target is computed post hoc for analysis. In manipulation, invalid actions are reflected in execution feedback, such as
a path-generation failure. EMHO explicitly exposes this feedback during
replanning and requires the next plan to differ from the failed attempt.
Rather than retrying the same action, the planner can revise the target
location or approach direction before attempting the manipulation again.

\noindent\textbf{Valid-but-Ineffective Recovery.}
A different failure mode occurs when individual actions are valid but collectively make little progress. In navigation, the agent may repeatedly rotate or reverse direction while every action executes successfully, and in manipulation, it may reissue nearly identical plans or repeatedly attempt grasps at the same coordinates. Because no explicit invalid-action signal is available, the harness must infer from the recent trajectory that the current behavior is ineffective. EMHO introduces mechanisms that detect these patterns and require the next plan to change strategy. In navigation, sequences of at least five consecutive rotations become less frequent. In manipulation, identical replans decrease from approximately 21\% to 18\%, while repeated grasps at the same coordinates decrease from approximately 30\% to 26\%. These results suggest that EMHO helps the agent recognize ineffective behavior even when the individual actions are valid.

\subsection{Non-Recovery Edits}
\label{sec:qual_nonrecovery}

\noindent\textbf{Progress Monitoring.}
Beyond responding to failure, EMHO changes how the harness determines whether an ongoing trajectory is making meaningful progress. In navigation, the evolved harness compares successive observations and uses changes in the target's apparent position and scale, as well as changes in scene appearance, to assess whether recent motion is bringing the agent closer to the target. In manipulation, as shown in Figure~\ref{fig:qual_non_recovery}, the evolved harness tracks execution phases, including approach, grasp, lift, translate, and release. During replanning, it uses the current gripper and object state to infer which phases have already been completed and which action should follow, rather than restarting the entire sequence. Consistent with this change, replanning traces more frequently identify the current execution phase, reaching approximately 21\% in some subtasks.

\noindent\textbf{Tool Selection and Orchestration.}
EMHO also changes how and when external vision tools are selected. The evolved harness selects tools based on the type of uncertainty relevant to the next decision. In navigation, depth estimation supports obstacle and proximity reasoning, while segmentation helps identify specific targets, and zoom provides finer visual details once the relevant region is identified. After blocked moves, the use of depth estimation rises from approximately 70\% to nearly 100\%, alongside an increase in the fraction of valid subsequent actions from approximately 44\% to 68\%. In manipulation, the harness learns to call depth estimation proactively for height-sensitive interactions, such as stacking and container placement, where relative height and clearance are important for subsequent actions.

\noindent\textbf{Perception and Visual Grounding.}
Finally, EMHO strengthens the connection between perceptual evidence and the actions that follow from it. In navigation, the evolved harness compares candidate targets using instruction-relevant attributes and uses the selected visual evidence to guide the subsequent motion decision. When the agent explicitly identifies a target in its reasoning, the arrival rate increases, and cases in which the agent passes the target and moves farther away decrease. In manipulation, the harness more explicitly distinguishes object identity from executable grasp coordinates and re-evaluates both from the current observation during planning. The fraction of scene descriptions listing at least three objects increases from approximately 8\% to 22\%, while incorrect object selection in initial plans decreases from 50\% to 34\% in spatial-relation tasks. These results indicate more detailed scene descriptions and fewer object-selection errors during planning.

\subsection{Evolution Across Optimization Iterations}
\label{sec:qual_trends}

\noindent\textbf{Recovery edits dominate early, while non-recovery edits emerge over time.}
Across the harness optimization process, the earliest proposals predominantly introduce recovery edits. Non-recovery edits are present from the start but tend to gain share only after the harness has acquired an explicit response to failure. This temporal pattern suggests that observable failure signals provide the most direct initial optimization target. In contrast, non-recovery edits require a more developed diagnosis of the ongoing trajectory, such as whether the embodied agent is making sufficient progress or whether its current plan should be adjusted before an explicit failure occurs. A detailed analysis of how the relative proportions of recovery and non-recovery edits change across optimization iterations is provided in Appendix~\ref{appendix:evolution_across_iterations}.

\noindent\textbf{Rules become more detailed.}
Later generations do not simply replace the first recovery rule; instead, they refine it by introducing increasingly detailed conditions. A representative navigation chain progresses from ``do not repeat the blocked action,'' to ``rotate after several blocked directions,'' and eventually to selecting a response based on the blocked direction, target heading, and the distance required to clear the obstacle. Across runs, recovery rules become more detailed over successive generations, with additional conditions and clauses that specify when and how each response should be applied. This suggests that the optimizer is not merely accumulating advice, but gradually transforming generic reactions into more detailed and context-dependent policies.

\section{Limitations and Future Work}
\label{sec:limitations}

As optimization proceeds, the harness that performs best on the search split is not always the one that performs best on held-out tasks. In several runs, later iterations continue to improve or maintain search performance while held-out performance plateaus or declines. This divergence suggests an emerging overfitting pattern, where the harness becomes increasingly specialized to the trajectories observed during optimization. These results motivate the need to consider generalization when selecting among harness iterations.

\section{Conclusion}

We introduced \textsc{EMHO}, a framework for improving embodied agents by evolving the harness around a frozen underlying model from experience traces and sparse environmental feedback. Rather than restricting adaptation to reusable skills or recovery prompts, \textsc{EMHO} optimizes the broader interaction procedure, including planning, context construction, and tool use. Across navigation and manipulation, EMHO improves search-set performance, and the resulting harnesses also outperform the initial harness on held-out tasks, suggesting that harness evolution can help translate existing model capabilities into more successful embodied behavior.
Our analysis shows that the evolved harness changes not only how agents recover from failures, but also how they monitor progress, gather perceptual evidence, and translate observations into subsequent actions. We further introduced EMHO-Merge, which uses episode-level gains and losses to refine a shared harness across subtasks and mitigate trade-offs during joint optimization. Together, these results show that harness optimization provides a promising direction for improving embodied-agent behavior without updating the underlying model.

\subsection*{AI use statement}

We used generative AI tools to aid and polish the writing and code development of this paper. We have not used generative AI tools for research ideation or execution, and generating synthetic datasets or proving mathematical claims are not applicable to this work. We take responsibility for the final content of this work, including text, claims, or artifacts produced with the aid of generative AI.

\subsection*{Acknowledgments}

This work was supported in part by NSF (IIS-2404180), IBM, and grants from the Institute of Information \& communications Technology Planning \& Evaluation (IITP) funded by the Korea government (MSIT) (No. 2022-0-00871, Development of AI Autonomy and Knowledge Enhancement for AI Agent Collaboration; No. RS-2025-2543949, Environment-Aware and Domain-Adaptive Multimodal Embodied AI for Real-World Interaction; No. RS-2019-II190079, Artificial Intelligence Graduate School Program (Korea University); and No. RS-2025-25439490). Additional support was provided by the Electronics and Telecommunications Research Institute (ETRI) grant (26CB1200, Development and Application of Science-Specialized Multimodal Foundation Models), the KOCCA grant (RS-2024-00345025), and the National Research Foundation of Korea (NRF) grant funded by the Korea government (MSIT) (RS-2025-25302986). This work also used resources provided through CIS251382 from the Advanced Cyberinfrastructure Coordination Ecosystem: Services \& Support (ACCESS) program, supported by NSF awards OAC-2138259, OAC-2138286, OAC-2138307, OAC-2137603, and OAC-2138296.

\setlength{\bibsep}{5pt}
\bibliography{references}
\bibliographystyle{references}

\newpage
\appendix

\section*{Appendix}

\section{Details of Harness Structure}
\label{appendix:harness_structure}

\begin{table}[h]
\centering
\caption{Overview of the four main functional components of the manipulation harness.}
\label{tab:harness_components}

\footnotesize
\setlength{\tabcolsep}{3.5pt}
\renewcommand{\arraystretch}{1.12}

\begin{tabularx}{\columnwidth}{
    @{}
    >{\raggedright\arraybackslash}p{0.20\columnwidth}
    >{\raggedright\arraybackslash}p{0.31\columnwidth}
    >{\raggedright\arraybackslash}X
    @{}
}
\toprule
\textbf{Component} &
\textbf{Main files} &
\textbf{Role in the harness} \\
\midrule

Planning &
\makecell[tl]{
    \texttt{SKILL.md} \\
    \texttt{system.md} \\
    \texttt{final\_planner.md} \\
    \texttt{agent.py} \\
    \texttt{planner.py} \\
    \texttt{model\_client.py}
} &
Defines the manipulation-planner contract and implements the main
decision loop, including model-input construction, plan generation,
action execution, and replanning from environment feedback.
\\[5pt]

Vision and tool use &
\makecell[tl]{
    \texttt{tool\_orchestrator.py} \\
    \texttt{vision\_policy.py} \\
    \texttt{tool\_select.md}
} &
Controls when and how visual evidence tools, such as zoom,
segmentation, and depth, are invoked and determines how their outputs
are incorporated into subsequent planning.
\\[5pt]

Memory and context &
\makecell[tl]{
    \texttt{module\_selector.py} \\
    \texttt{memory.py} \\
    \texttt{memory\_render.md} \\
    \texttt{safe\_context.md}
} &
Maintains bounded episode history and selects the model-facing context,
including previous actions, observations, feedback, and other safe
contextual information used during replanning.
\\[5pt]

Validation &
\makecell[tl]{
    \texttt{parser.py} \\
    \texttt{repair.py} \\
    \texttt{repair.md}
} &
Validates generated plans and actions, corrects malformed or
non-executable outputs, and ensures that the resulting plan is executable.
\\

\bottomrule
\end{tabularx}
\end{table}

The harness consists of the instructions and runtime components that control planning, context construction, tool use, action generation, validation, and recovery. During optimization, EMHO modifies these components based on recurring failure patterns observed in experience traces.

\paragraph{EMHO Harness Optimizer Instruction Example}
This example illustrates the instruction provided to the harness optimizer for the Navigation Base subtask. It shows the general structure of the optimizer instruction, including the available evidence, editable harness components, regression-aware constraints, and required workflow. File and path names are shown in simplified placeholder form for presentation.

%
%
%
%

\par
\vspace{1em}

\begingroup

\setlength{\fboxsep}{0pt}
\setlength{\parindent}{0pt}
\setlength{\parskip}{0pt}

\definecolor{PromptInk}{HTML}{152A4A}
\definecolor{PromptHeaderBG}{HTML}{DDE1E6}

\newlength{\PromptSourceWidth}
\setlength{\PromptSourceWidth}{612.375pt}

\newcommand{\PromptMono}{\ttfamily}
\newcommand{\PromptMonoBold}{\ttfamily\bfseries}

\newcommand{\PromptScaleRow}[1]{%
  \noindent
  \resizebox{\linewidth}{!}{#1}\par%
}

\newcommand{\PromptHeader}[1]{%
  \PromptScaleRow{%
    \colorbox{PromptHeaderBG}{%
      \makebox[\PromptSourceWidth][l]{%
        \hspace*{8pt}%
        \rule{0pt}{16.5pt}%
        \raisebox{0.8pt}{%
          \PromptMonoBold
          \fontsize{13pt}{13pt}\selectfont
          \color{PromptInk}%
          #1%
        }%
      }%
    }%
  }%
}

\newcommand{\PromptLine}[2][0]{%
  \PromptScaleRow{%
    \colorbox{white}{%
      \makebox[\PromptSourceWidth][l]{%
        \rule{0pt}{12.5pt}%
        \hspace*{8pt}%
        \hspace*{#1pt}%
        \raisebox{0.1pt}{%
          \PromptMono
          \fontsize{9.5pt}{11.5pt}\selectfont
          \color{PromptInk}%
          #2%
        }%
      }%
    }%
  }%
}

\newcommand{\PromptGap}[1]{%
  \par\vspace*{#1pt}%
}

\offinterlineskip

\PromptHeader{EMHO  Harness optimizer instruction example}

\PromptGap{6}

\PromptLine{Runtime workspace paths:}

\PromptGap{5}

\PromptLine{- read-only proposer\_view root: \{\{PROPOSER\_VIEW\_ABSOLUTE\_PATH\}\}}
\PromptLine{- editable candidate root: \{\{CANDIDATE\_ABSOLUTE\_PATH\}\}}

\PromptGap{8}

\PromptLine{You are optimizing the EmbodiedBench navigation base subtask harness to improve task success.}

\PromptGap{8}

\PromptLine{Use the visible evidence in \{\{file\_path\}\}, and the current harness snapshot, to find}
\PromptLine{harness changes that are likely to improve navigation performance.}

\PromptGap{8}

\PromptLine{Make focused, evidence-supported changes to the allowed candidate harness files.}
\PromptLine{The changes should improve how the agent selects episode memory or vision}
\PromptLine{support, formats outputs, repairs mistakes, or plans actions.}

\PromptGap{8}

\PromptLine{You must make at least one substantive change to an allowed candidate harness}
\PromptLine{file. An unchanged copy of the parent harness is rejected as a failed proposal.}
\PromptLine{Keep the patch practical and coherent. Edit the files needed for the improvement.}
\PromptLine{The goal is better task success, not producing the smallest possible diff.}

\PromptGap{8}

\PromptLine{Changes must be general and non-episode-specific. Do not use hidden simulator}
\PromptLine{state, shortest paths, heldout data, or hard-coded episode solutions. Do not force}
\PromptLine{narrow tool/action recipes; improve the harness behavior so the model can make better}
\PromptLine{decisions. Do not stop after describing a plan or saying that you will read files.}
\PromptLine{Make the edits, then write \{\{file\_path\}\} with cited evidence}
\PromptLine{and changed files.}

\PromptGap{10}

\PromptLine{Read only:}

\PromptGap{4}

\PromptLine{- \{\{PROPOSER\_VIEW\_ABSOLUTE\_PATH\}\}}

\PromptGap{10}

\PromptLine{Edit only these candidate workspace harness files: \{\{file\_paths\}\}}

\PromptGap{8}

\PromptLine{Make a coherent set of changes to the allowed harness files when needed.}
\PromptLine{Prefer the coordinated update that best supports task success.}

\PromptGap{10}

\PromptLine{Regression-aware editing:}

\PromptGap{5}

\PromptLine{- Review candidate scores, parent candidates, changed harness paths, score deltas,}
\PromptLine[8]{and harness\_diff\_from\_parent. If a lower-scoring child added or strengthened a rule,}
\PromptLine[8]{tool-selection policy, prompt section, or parser behavior, do not repeat that}
\PromptLine[8]{structure unless the evidence clearly explains why your version avoids the regression.}

\PromptGap{5}

\PromptLine{- Before adding a new rule, ask whether an existing rule should be removed,}
\PromptLine[8]{shortened, weakened, or reverted. A valid candidate may improve by deleting or}
\PromptLine[8]{simplifying text.}

\PromptGap{10}

\PromptLine{Rules:}

\PromptGap{5}

\PromptLine{- Do not use heldout data, hidden simulator state, coordinates, rewards, shortest}
\PromptLine[8]{paths, or episode-specific hacks.}

\PromptLine{- Do not edit evaluator, dataset splits, datasets, runner scripts, or action semantics.}

\PromptLine{- Preserve CandidateHarness API.}

\PromptLine{- Cite search evidence from search\_evidence.jsonl in rationale using event\_id,}
\PromptLine[8]{episode\_ref, env\_step, or record\_type where available.}

\PromptGap{10}

\PromptLine{Workflow:}

\PromptGap{5}

\PromptLine{1. Read \{\{file\_path\}\} and any relevant current prompts.}

\PromptLine{2. If \{\{file\_path\}\} exists, compare it only as optional context because}
\PromptLine[8]{it differs from source\_harness.}

\PromptLine{3. Check frontier.json for parent-success episodes lost by later candidates and for}
\PromptLine[8]{harness changes correlated with lower scores.}

\PromptLine{4. Review the visible search evidence and regression evidence for harness changes}
\PromptLine[8]{likely to improve task success.}

\PromptLine{5. Edit the allowed files needed for a coherent improvement.}

\PromptLine{6. Write \{\{file\_path\}\} with cited evidence, changed files, and why}
\PromptLine[8]{the patch should improve task success.}

\PromptLine{7. Run \{\{file\_path\}\} only for syntax check. }

\endgroup

\paragraph{EMHO-Merge Harness Optimizer Instruction Example}
This example illustrates the instruction provided to the EMHO-Merge harness optimizer for the Navigation Base subtask.  
File and path names are shown in simplified placeholder form for presentation.

%
%
%
%

\par
\vspace{1em}

\begingroup

\setlength{\fboxsep}{0pt}
\setlength{\parindent}{0pt}
\setlength{\parskip}{0pt}

\definecolor{PromptInk}{HTML}{152A4A}
\definecolor{PromptHeaderBG}{HTML}{DDE1E6}

\setlength{\PromptSourceWidth}{612.375pt}

\newcommand{\PromptMono}{\ttfamily}
\newcommand{\PromptMonoBold}{\ttfamily\bfseries}

\newcommand{\PromptScaleRow}[1]{%
  \noindent
  \resizebox{\linewidth}{!}{#1}\par%
}

\newcommand{\PromptHeader}[1]{%
  \PromptScaleRow{%
    \colorbox{PromptHeaderBG}{%
      \makebox[\PromptSourceWidth][l]{%
        \hspace*{8pt}%
        \rule{0pt}{16.5pt}%
        \raisebox{0.8pt}{%
          \PromptMonoBold
          \fontsize{13pt}{13pt}\selectfont
          \color{PromptInk}%
          #1%
        }%
      }%
    }%
  }%
}

\newcommand{\PromptLine}[2][0]{%
  \PromptScaleRow{%
    \colorbox{white}{%
      \makebox[\PromptSourceWidth][l]{%
        \rule{0pt}{12.5pt}%
        \hspace*{8pt}%
        \hspace*{#1pt}%
        \raisebox{0.1pt}{%
          \PromptMono
          \fontsize{9.5pt}{11.5pt}\selectfont
          \color{PromptInk}%
          #2%
        }%
      }%
    }%
  }%
}

\newcommand{\PromptGap}[1]{%
  \par\vspace*{#1pt}%
}

\offinterlineskip

\PromptHeader{EMHO-Merge Harness optimizer instruction example}

\PromptGap{6}

\PromptLine{Runtime workspace paths:}

\PromptGap{5}

\PromptLine{- read-only proposer\_view root: \{\{PROPOSER\_VIEW\_ABSOLUTE\_PATH\}\}}
\PromptLine{- editable candidate root: \{\{CANDIDATE\_ABSOLUTE\_PATH\}\}}

\PromptGap{8}

\PromptLine{You are optimizing the EmbodiedBench navigation base subtask harness to improve}
\PromptLine{task success.}

\PromptGap{8}

\PromptLine{Use the visible evidence in \{\{file\_path\}\}, and the current harness snapshot, to find}
\PromptLine{harness changes that are likely to improve navigation performance.}

\PromptGap{8}

\PromptLine{Make focused, evidence-supported changes to the allowed candidate harness files.}
\PromptLine{The changes should improve how the agent selects episode memory or vision}
\PromptLine{support, formats outputs, repairs mistakes, or plans actions.}

\PromptGap{8}

\PromptLine{Follow the proposal mode supplied by the optimizer. If the mode is `new\_mechanism`,}
\PromptLine{propose a genuinely new behavior. If the mode is `repair\_or\_new`, investigate the}
\PromptLine{previous change using the required paired trajectories before deciding between}
\PromptLine{`repair` and `new`. Record the selected mode and its supporting evidence in}
\PromptLine{`experiment.json`.}

\PromptGap{8}

\PromptLine{- `new\_mechanism`: test a genuinely different behavior, not a reworded version of}
\PromptLine[8]{any mechanism in `forbidden\_new\_mechanism\_ids`.}

\PromptGap{5}

\PromptLine{- `repair\_or\_new`: the latest evaluated change produced both gained and lost}
\PromptLine[8]{episodes. This allows one repair investigation but does not establish causality.}
\PromptLine[8]{Inspect both parent and candidate trajectories for at least one gained and one}
\PromptLine[8]{lost episode from any active subset. Verify that the changed behavior appears and}
\PromptLine[8]{identify a visible condition that separates benefit from harm. Choose `repair`}
\PromptLine[8]{only when the evidence supports that condition; otherwise choose `new` and}
\PromptLine[8]{explain `why\_not\_repair`.}

\PromptGap{5}

\PromptLine{- A repair must reuse the original `mechanism\_id` and specify `repair\_of`. Each}
\PromptLine[8]{mechanism receives only one evaluated repair attempt. Mechanical validation}
\PromptLine[8]{retries do not count.}

\PromptGap{5}

\PromptLine{- The writable candidate starts from the current best parent, which may differ from}
\PromptLine[8]{the previous candidate. Interpret `previous\_change.diff` relative to its actual}
\PromptLine[8]{parent. Rejected changes are not installed. A repair may reconstruct only the}
\PromptLine[8]{useful behavior with a narrower trigger.}

\PromptGap{5}

\PromptLine{- Treat small differences as uncertain. Do not invent causal explanations or assume}
\PromptLine[8]{additional evaluation or parallel proposal budgets.}

\PromptGap{5}

\PromptLine{- For both `new` and `repair`, inspect and cite the parent and candidate event IDs}
\PromptLine[8]{for every category in `required\_previous\_pair\_categories`. Use}
\PromptLine[8]{`comparison -{}-category gained` and `comparison -{}-category lost` to find eligible}
\PromptLine[8]{IDs, then query each trajectory before writing the contract. Baseline-only}
\PromptLine[8]{examples do not satisfy this requirement.}

\PromptGap{5}

\PromptLine{- Novelty is checked against behavior descriptions and patch content, including}
\PromptLine[8]{`discouraged\_behavior\_families` and `previous\_patch\_signatures`. Renamed rotation,}
\PromptLine[8]{blocked-move, detour, lateral-step, or reposition warnings still belong to the}
\PromptLine[8]{same recovery family. They cannot be submitted as new mechanisms, although one}
\PromptLine[8]{evidence-supported repair is allowed.}

\PromptGap{5}

\PromptLine{- Before adding an intervention, inspect a successful trajectory where the proposed}
\PromptLine[8]{trigger also occurs. Compare the visible feedback and subsequent progress. If}
\PromptLine[8]{harmful and useful activations cannot be distinguished, narrow or abandon the}
\PromptLine[8]{intervention. Raw action counts alone do not prove the agent is stuck. Record the}
\PromptLine[8]{counterexample and remaining uncertainty in `regression\_risk`.}

\PromptGap{8}

\PromptLine{For a repair also supply `repair\_of` and `separating\_evidence` (including the observed}
\PromptLine{activation and the visible distinction between gained/lost situations). Both proposal}
\PromptLine{modes must cite BOTH parent and candidate event IDs for each required gained/lost}
\PromptLine{category; the subtask's net score sign is irrelevant. For a new mechanism when a}
\PromptLine{repair was offered, supply `why\_not\_repair`. Use a stable `mechanism\_id` across}
\PromptLine{repairs; renaming the same warning is not a new mechanism. This metadata is}
\PromptLine{optimizer-only and must not enter runtime prompts.}

\PromptGap{10}

\PromptLine{Workflow:}

\PromptGap{5}

\PromptLine{1. Read \{\{file\_path\}\}. Read the relevant hunks of `previous\_change.diff` and the}
\PromptLine[8]{hypothesis/change sections of \{\{file\_path\}\} when available.}

\PromptGap{4}

\PromptLine{2. Briefly retain the observed failure/success differences across subsets as you go.}

\PromptGap{4}

\PromptLine{3. Now select one mechanism supported by that review. The completed review queue}
\PromptLine[8]{includes required previous gained/lost pairs. Use `comparison` to identify their}
\PromptLine[8]{category and cite BOTH event IDs per required category.}

\PromptGap{4}

\PromptLine{4. Explain both prospective gains and regressions. State which subsets the mechanism}
\PromptLine[8]{affects, which evidence contradicts it, and what the previous edit taught you. A}
\PromptLine[8]{lower score alone does not prove an idea failed: check implementation activation}
\PromptLine[8]{and run variability. Do not repeat a prior unsuccessful change without a specific}
\PromptLine[8]{correction or new evidence. Keep overall search success as the objective; do not}
\PromptLine[8]{require every subset to improve in every sample.}

\PromptGap{4}

\PromptLine{5. Read then write the draft `experiment.json` before further source research. The}
\PromptLine[8]{draft is editable until the first harness edit; writing it does not waive novelty}
\PromptLine[8]{or final validation. Once you choose the hypothesis, read only the details needed}
\PromptLine[8]{to implement it. Use \{\{file\_path\}\} to locate exact source ranges.}

\PromptGap{4}

\PromptLine{6. Make one coherent edit in the existing allowlisted files.}

\PromptGap{4}

\PromptLine{7. Record what was implemented, checks performed, and unresolved assumptions; do}
\PromptLine[8]{not claim an unrun experiment succeeded.}

\endgroup

\newpage 
\section{Quantitative Results}
\label{appendix:quantitative_result}

This appendix presents full quantitative results for all navigation and
manipulation subtasks. For each subtask, the plots show the best success
rate attained separately on the search and held-out splits up to each
optimization iteration. In each EMHO run, the same model serves as both
the embodied agent and the harness optimizer. Baseline results for other
underlying models are taken from Embodied Tool Protocol~\citep{embodiedtool}
and EmbodiedBench~\citep{embodiedbench}.
Throughout optimization, candidate generation and parent updates are guided by search-set feedback and harness history; held-out trajectories and scores are not provided to the optimizer. The split-wise maxima characterize the best observed performance among the resulting candidates, complementing our analysis of how experience-guided revisions change the harness and agent behavior. Since the search and held-out maxima may occur at different iterations, we interpret the held-out results as retrospective summaries of candidate performance, rather than estimates of the generalization performance of a single search-selected harness.

\begin{figure}[ht]
    \centering
    \includegraphics[width=0.9\linewidth]{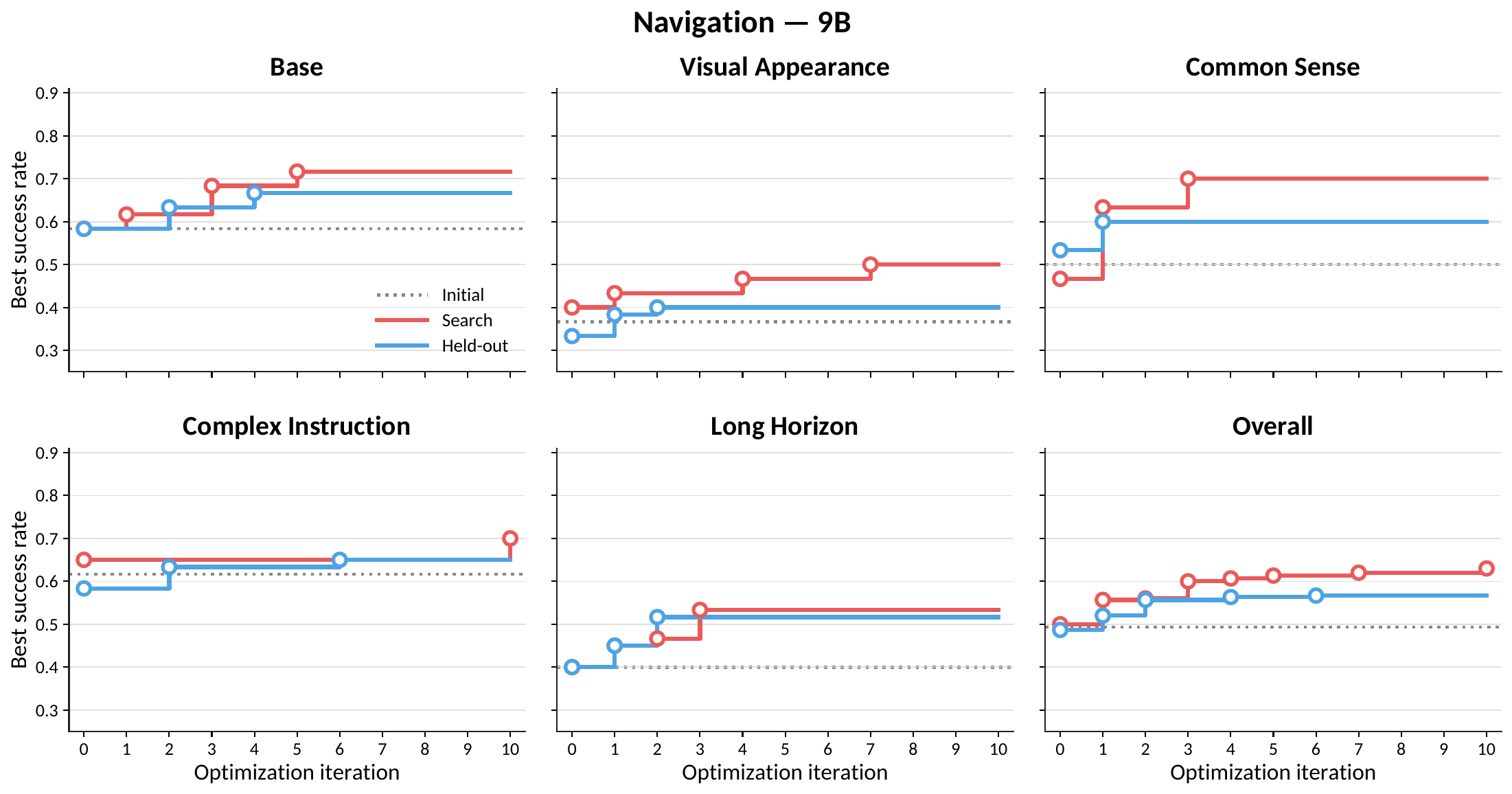}
    \caption{Quantitative results on the navigation task with the 9B model across optimization iterations.}
    \label{fig:app-main-figure-navigation-9b}
\end{figure}

\begin{figure}[ht]
    \centering
    \includegraphics[width=0.9\linewidth]{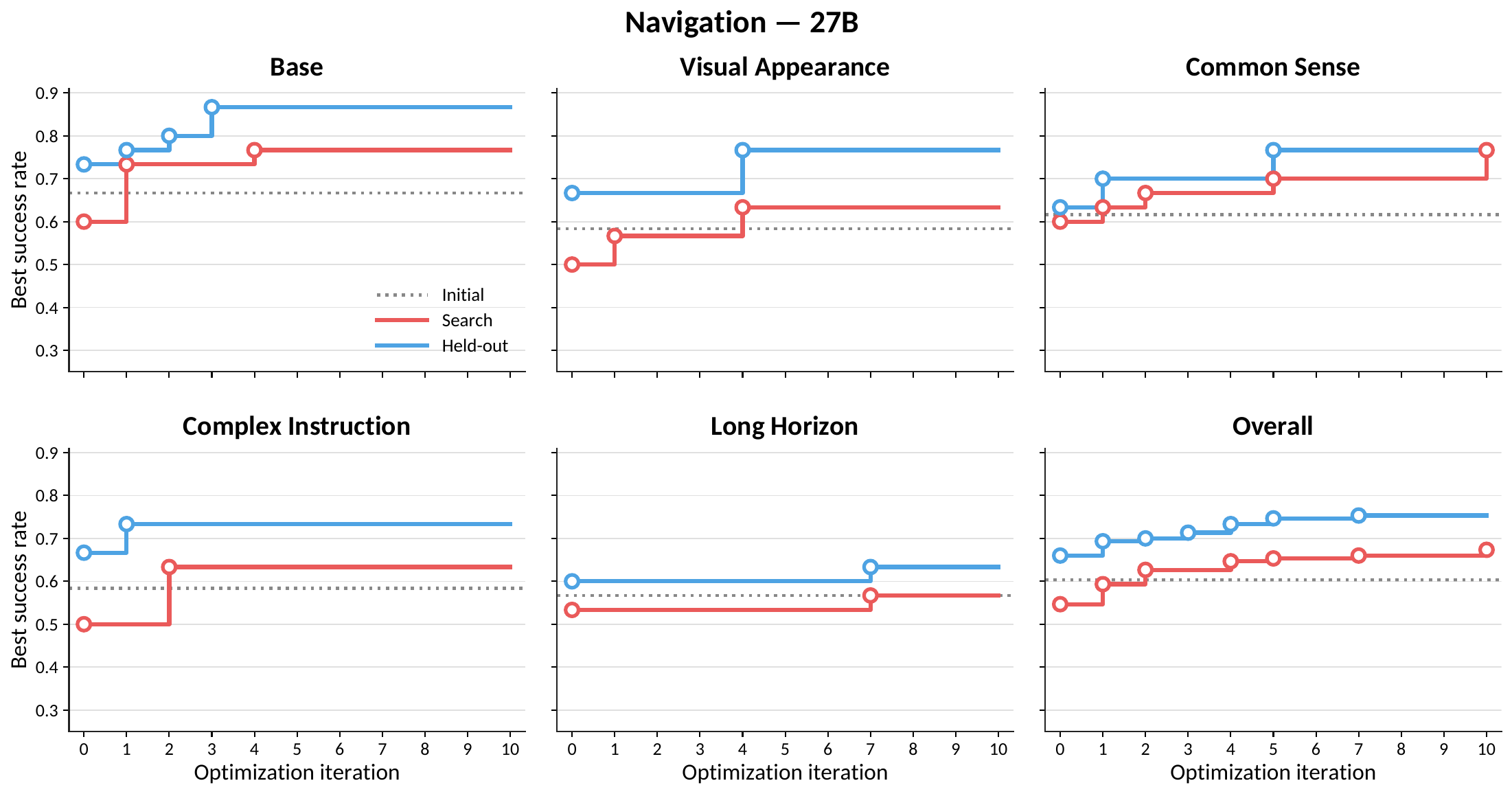}
    \caption{Quantitative results on the navigation task with the 27B model across optimization iterations.}
    \label{fig:app-main-figure-navigation-27b}
\end{figure}

\begin{figure}[ht]
    \centering
    \includegraphics[width=0.9\linewidth]{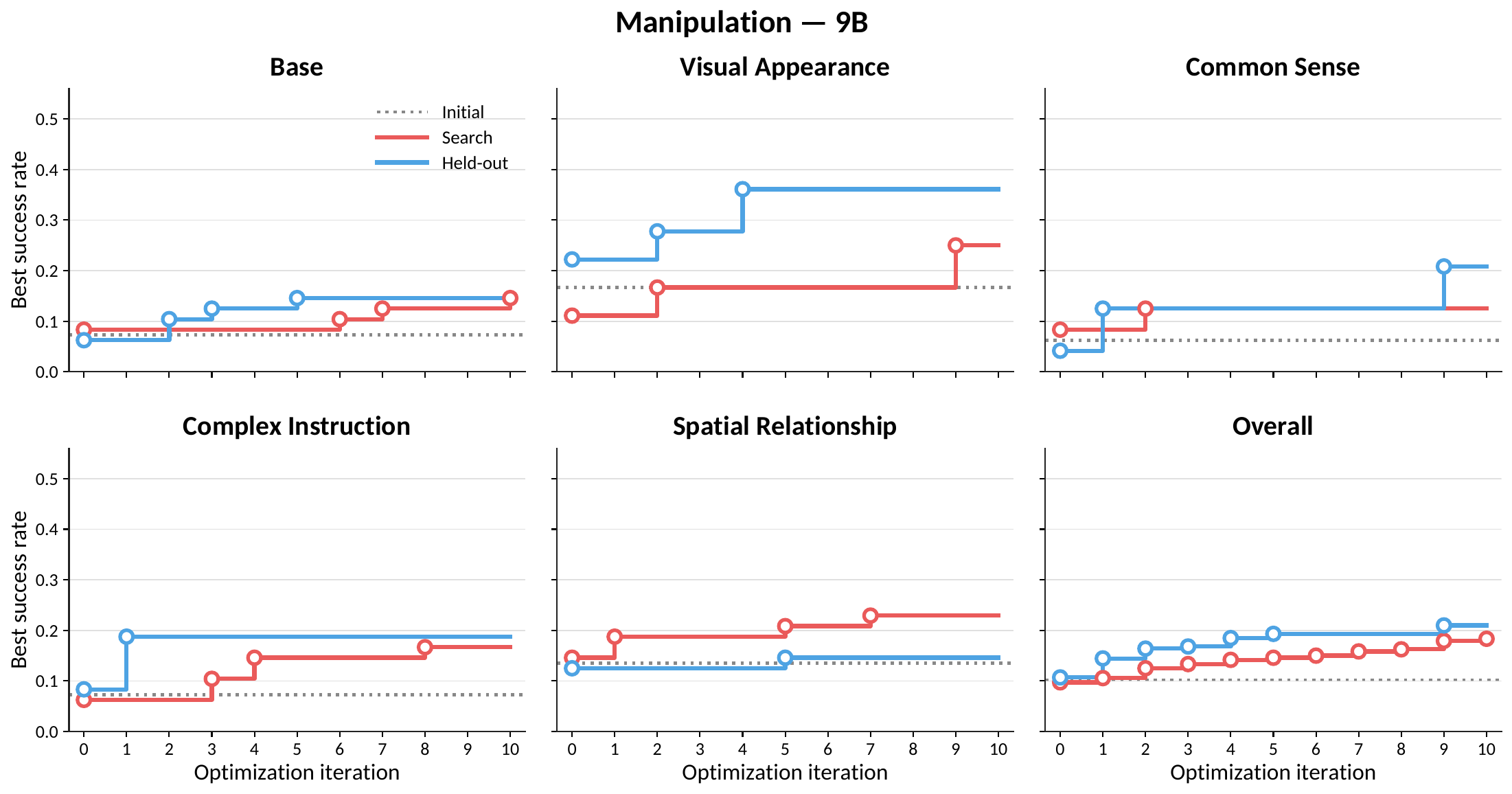}
    \caption{Quantitative results on the manipulation task with the 9B model across optimization iterations.}
    \label{fig:app-main-figure-manipulation-9b}
\end{figure}

\begin{figure}[ht]
    \centering
    \includegraphics[width=0.9\linewidth]{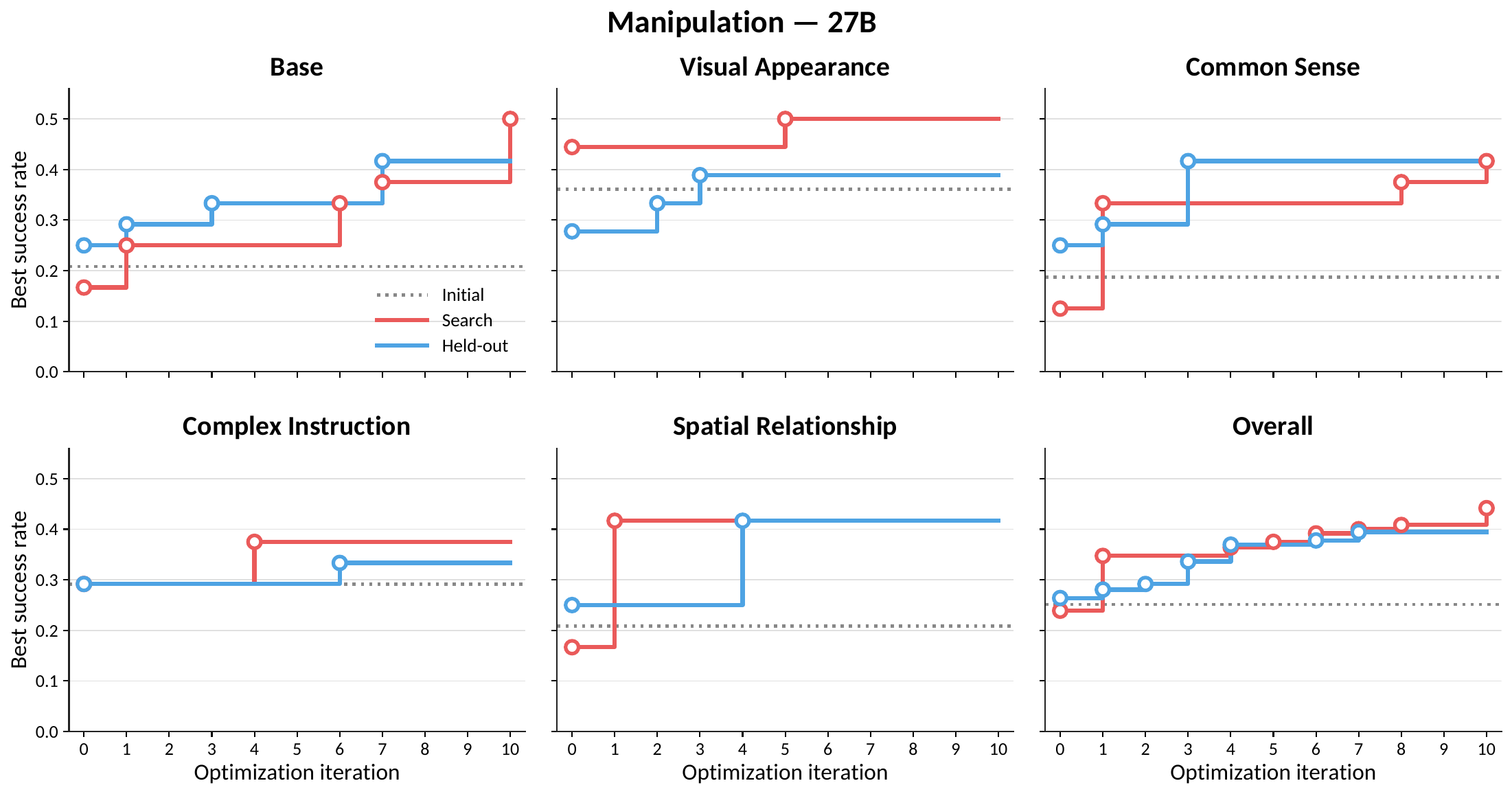}
    \caption{Quantitative results on the manipulation task with the 27B model across optimization iterations.}
    \label{fig:app-main-figure-manipulation-27b}
\end{figure}

\newpage

\section{Ablation Study}
\label{appendix:history_ablation}

EMHO provides the harness optimizer with two forms of historical context: episode-level experience traces, $\mathcal{E}^{(0:i)}$, and the history of previous harness revisions and their evaluation outcomes, $\mathbf{H}_{\mathrm{hist}}^{(0:i)}$. We ablate these two sources of information on the Navigation Base subtask to examine their respective contributions. As shown in Table~\ref{tab:optimization_context_ablation}, Full EMHO receives both experience traces and harness history. The  `without Experience Traces' variant removes episode-level trajectories while retaining previous harness revisions and their scores; the `without Harness History' variant retains the trajectories but hides prior harness modifications and candidate history. Finally, `Score Only' removes both forms of historical context and provides only the current parent harness and its evaluation score. The comparison isolates the complementary roles of the two information sources. Experience traces provide behavioral evidence about \emph{why} an evaluated harness succeeds or fails, while harness history informs the optimizer about \emph{what} modifications have already been attempted and how they affected performance. Comparing the full setting against the two single-source ablations therefore measures whether either source alone is sufficient, while the score-only setting tests whether scalar evaluation feedback can guide harness optimization without trajectory-level or revision-level context.

Full EMHO achieves the strongest performance on the search set, while removing either source of context degrades performance. This suggests that the two histories provide complementary information: traces help diagnose recurring behavioral failures, whereas revision history helps the optimizer avoid revisiting ineffective modifications and refine previously useful ones. 
\begin{table}[t]
\centering
\caption{
Ablation of the optimization context on the Navigation Base subtask search set.
All variants use the same model and evaluation procedure, while differing
only in whether experience traces and harness-revision history are provided.
Performance is reported on the search set.
}
\label{tab:optimization_context_ablation}
\small
\setlength{\tabcolsep}{6pt}
\begin{tabular}{lccc}
\toprule
\textbf{Method}
& \textbf{Traces}
& \textbf{History}
& \textbf{Search} \\
\midrule
Full EMHO
& \checkmark
& \checkmark
& \textbf{76.7} \\

w/o Experience Traces
& --
& \checkmark
& 70.0 \\

w/o Harness History
& \checkmark
& --
& 66.7 \\

Score Only
& --
& --
& 63.3 \\
\bottomrule
\end{tabular}
\end{table}

\section{Qualitative Results}
\label{appendix:qualitative_results}

Aggregate task success reveals whether harness optimization works, but not \emph{what the optimizer changes}. We therefore compare the initial and evolved harnesses on held-out episodes. We organize the resulting edits into two groups. Recovery edits govern how the agent responds once a failure is observed or inferred, whereas non-recovery edits operate independently of failure signals, shaping how the agent processes observations, reasons about the current state, and selects actions during ongoing interaction. For each category, we first describe the harness change, then examine a representative episode, and finally measure whether the intended behavior appears across held-out trajectories.

\subsection{Recovery Edits}
\label{sec:app_qual_recovery}

EmbodiedBench provides only lightweight feedback about the outcome of interaction. In navigation, the environment reports action-level outcomes such as valid, invalid, or blocked. In manipulation, it provides action-validity feedback, a scalar reward in ${-1,0,1}$, and whether the agent is currently holding an object. These signals expose limited information about execution outcomes, but generally do not identify the underlying cause of failure. We divide recovery edits into \textbf{invalid-action recovery}, which responds to actions reported as invalid or blocked, and \textbf{valid-but-ineffective action recovery}, which responds when actions execute validly but repeated attempts or unchanged observations show that the task is not progressing.

\begin{figure}[ht]
    \centering
    \includegraphics[width=\linewidth]{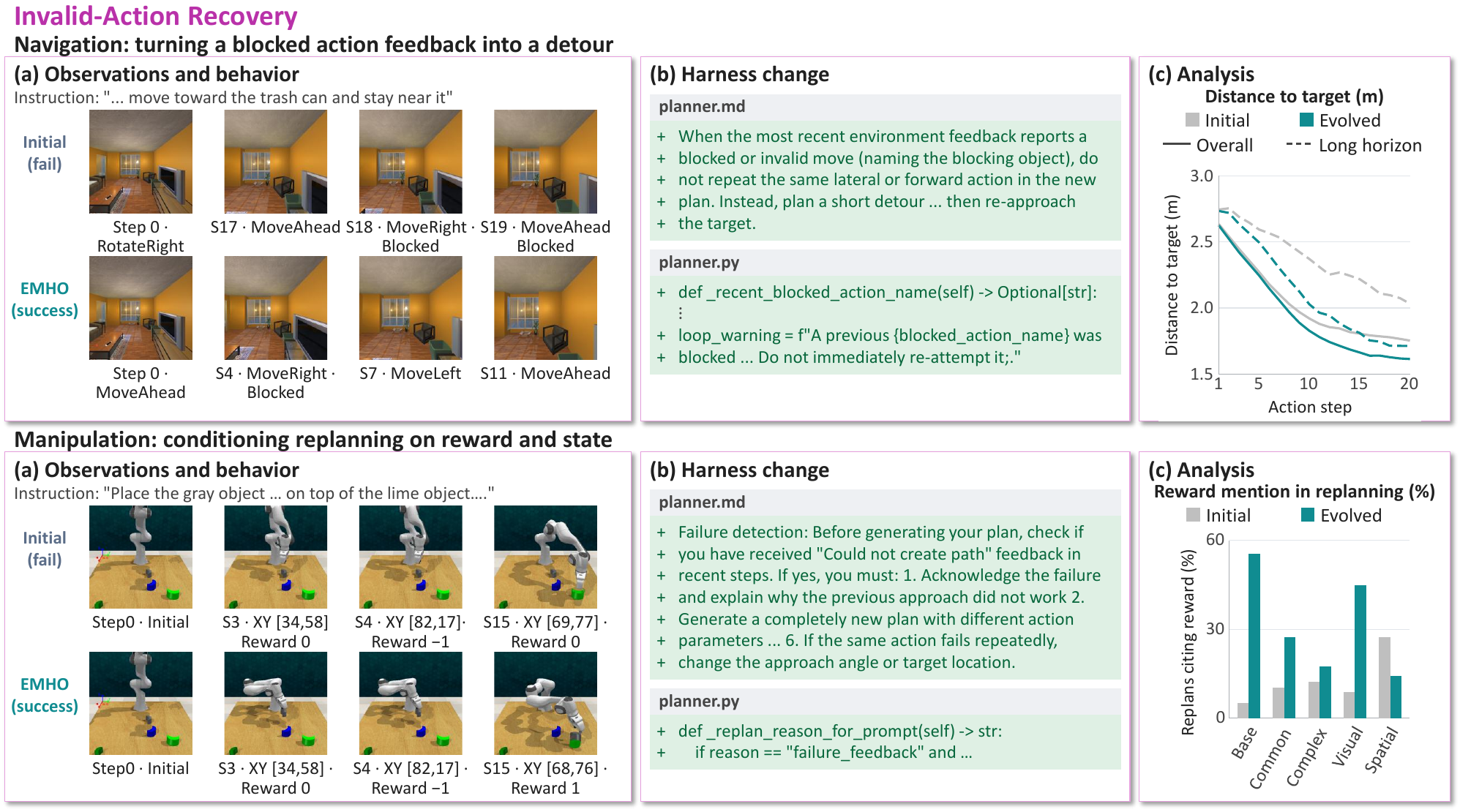}
    \caption{Qualitative analysis for invalid recovery in navigation and manipulation.}
    \label{fig:app-emho-qualitative-analysis-invalid-recovery}
\end{figure}

\noindent\textbf{Invalid-Action Recovery.}
\label{sec:qual_feedback_recovery}

\noindent\emph{Navigation: turning blocked-action feedback into a detour.}
When a navigation action is blocked, the initial harness often remains trapped around the same obstacle, proposing alternative actions without actually escaping the local constraint. EMHO evolves the harness to retain and act on this failure information during subsequent replanning. The resulting edits introduce explicit recovery rules that discourage movement along recently blocked directions and instead favor reorientation, backtracking, or lateral detours before re-approaching the target. They also add mechanisms that identify which action and direction were blocked, track multiple recent failures, and summarize these constraints for the planner, ensuring that this information remains available across replanning steps rather than being lost after the immediate feedback. In some cases, the blocked-state information is further carried through memory and tool-derived evidence, allowing later decisions to remain conditioned on what has already failed. Together, these changes turn blocked feedback from a transient execution message into a persistent constraint on subsequent planning. We assess the effect of this recovery behavior by tracking the agent's distance to the target over the course of each episode. This metric is used only for post-hoc analysis and is not provided to the optimizer during harness optimization. With the initial harness, trajectories often make limited progress toward the target and remain at a similar distance for multiple steps after encountering blockage. In contrast, evolved harnesses tend to continue reducing the distance to the target, with the clearest difference observed in long-horizon episodes.

\noindent\emph{Manipulation: conditioning replanning on reward and state.}
Manipulation provides limited feedback after plan execution: a reward of $-1$ indicates an invalid action, $0$ indicates valid execution without task completion, and $1$ indicates task success, together with a state signal indicating whether the object is currently held. EMHO evolves the harness to explicitly incorporate this feedback into replanning. In particular, when an action receives invalid feedback, the planner is instructed not to simply repeat the same attempt, but to revise the next plan by changing relevant action parameters, such as the target location or approach direction. Related edits also expose the most recent reward and execution outcome to the planner, so that the revised plan is conditioned on what failed rather than generated independently of the previous attempt. To quantify this behavior, we measure how often replanning traces explicitly refer to the observed reward feedback. This fraction increases most clearly for the base, common-sense, and visual task families, rising from approximately 5\% to 55\%, 10\% to 28\%, and 10\% to 45\%, respectively.

\begin{figure}[ht]
    \centering
    \includegraphics[width=\linewidth]{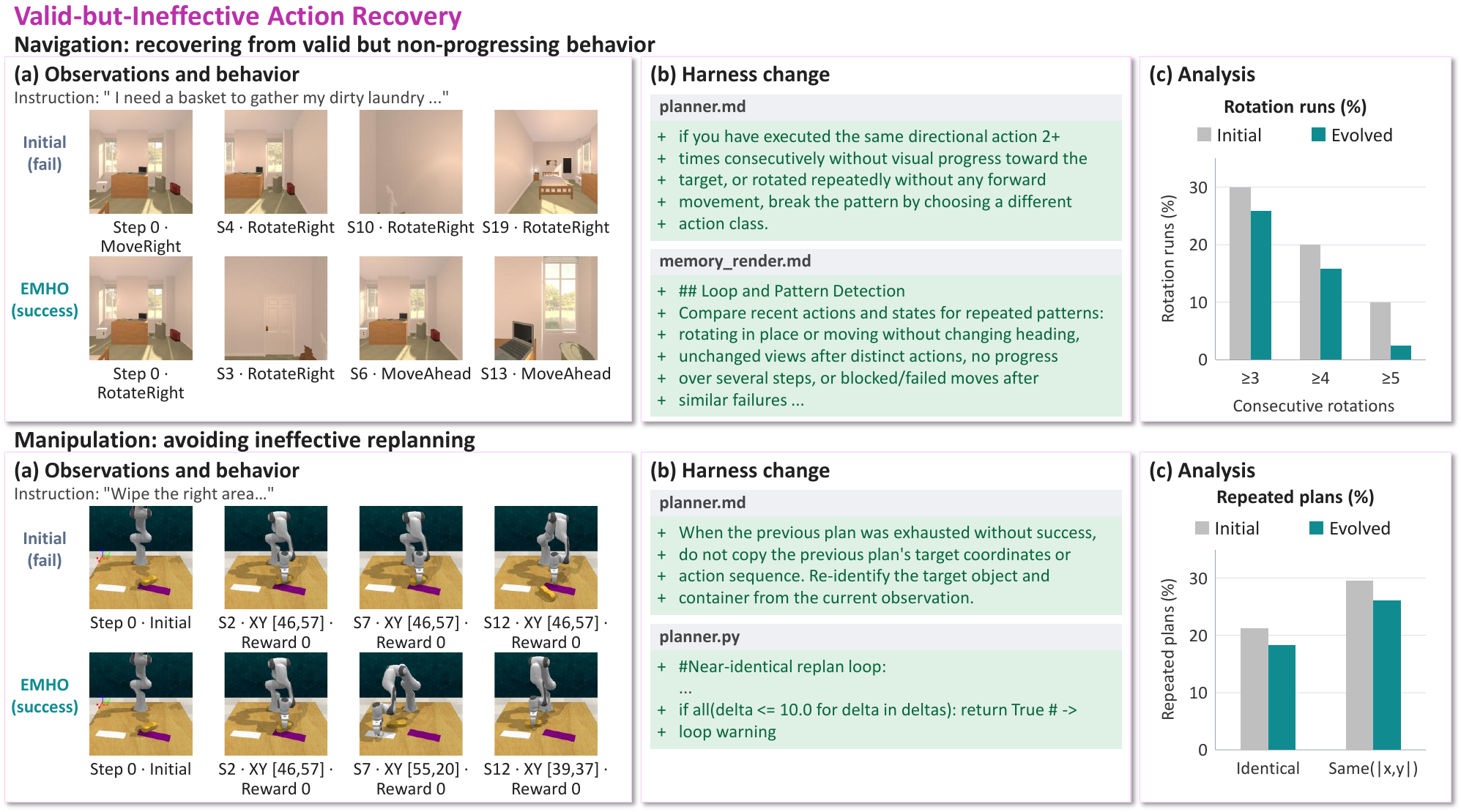}
    \caption{Qualitative analysis for valid but ineffective action recovery in navigation and manipulation.}
    \label{fig:app-emho-qualitative-analysis-valid-recovery}
\end{figure}

\noindent\textbf{Valid-but-Ineffective Action Recovery.}
\label{sec:qual_nonfeedback_recovery}


\noindent\emph{Navigation: recovering from valid but non-progressing behavior.} In navigation,  the agent can continue issuing valid actions while repeatedly rotating, reversing direction, or otherwise failing to make progress toward the target. EMHO introduces mechanisms that recognize such behavior from the recent interaction history and make it visible during replanning. The evolved harness summarizes repeated or reversing actions, compares recent observations to determine whether the agent is making meaningful progress, and encourages a change in action strategy when valid actions repeatedly leave the agent in a similar state. These changes help the planner distinguish successful execution from actual task progress.  We count runs of at least three, four, and five consecutive rotations in held-out trajectories. The evolved harness reduces repetition at all three thresholds, with runs of at least five rotations decreasing from approximately 10\% to 3\%.

\noindent\emph{Manipulation: avoiding ineffective replanning.}
A similar failure mode arises in manipulation when an executable plan completes without solving the task, yet replanning produces essentially the same plan or returns to the same grasp location. EMHO modifies the harness to retain what was attempted and require the subsequent plan to make a meaningful change, such as revising the grasp point, approach height, descent depth, or target estimate. Additional mechanisms compare consecutive plans for near-duplicate behavior and prompt the agent to re-localize the relevant object when replanning would otherwise repeat an ineffective attempt. We compare consecutive replans for identical action sequences and reuse of the same grasp coordinates. Both forms of repetition become less frequent with the evolved harness.

\begin{figure}[ht]
    \centering
    \includegraphics[width=\linewidth]{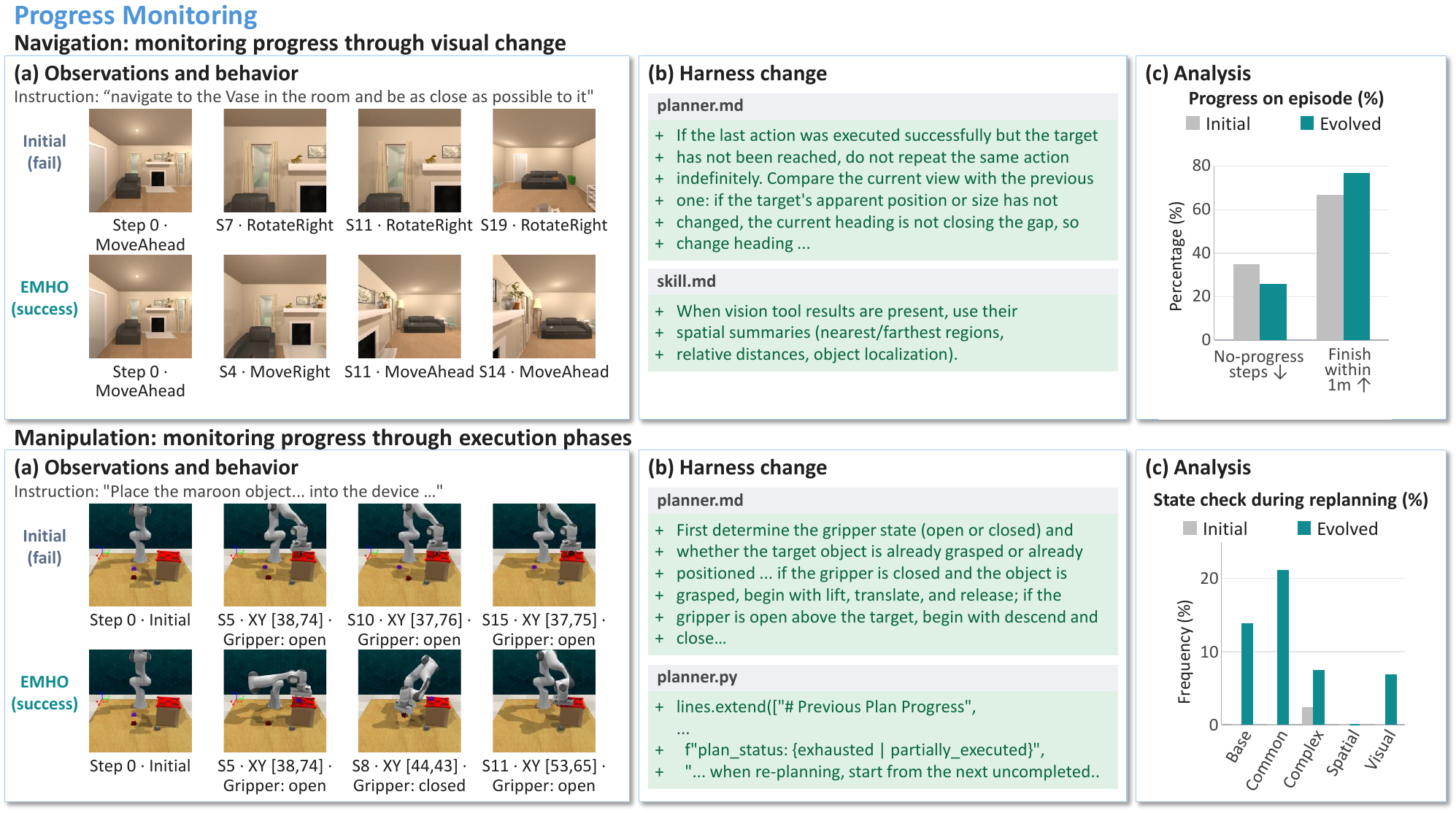}
    \caption{Qualitative analysis for progress monitoring in navigation and manipulation.}
    \label{fig:app-emho-qualitative-analysis-progress}
\end{figure}

\subsection{Non-Recovery Edits}
\label{sec:app_qual_nonrecovery}

\noindent\textbf{Progress Monitoring.}
\label{sec:qual_progress}

\noindent\emph{Navigation: monitoring progress through visual change.} EMHO evolves the navigation harness to explicitly monitor whether recent actions are producing observable progress toward the task. The harness compares consecutive observations after successful actions and uses changes in the target's apparent position, size, or surrounding scene to determine whether the current motion is advancing the agent or leaving it effectively stationary. When little visual progress is observed, the planner is encouraged to change its strategy---for example, by rotating, translating laterally, or moving to a new viewpoint---rather than continuing along the same heading. Related edits shorten cached multi-action plans so that the agent can re-observe the scene and reassess progress before committing to additional motion. Across held-out visual-appearance episodes, no-progress steps decrease from 35\% to 26\%, while the fraction of episodes ending within one meter of the target increases from 67\% to 77\%. These changes indicate that the evolved harness more consistently uses observable state changes to determine whether its current behavior is making meaningful progress.

\noindent\emph{Manipulation: monitoring progress through execution phases.} EMHO evolves the manipulation harness to organize execution into an explicit five-stage process: approach with the gripper open, descend and grasp, lift, translate, and release. The harness then introduces stage-specific behavioral rules that determine which phase has been completed, whether the expected state transition occurred, and what action should follow. Before replanning, the agent examines the current observation together with the gripper and object state to determine whether the object has been grasped, lifted, moved, or already positioned, and either continues from the next unfinished phase or returns to an earlier phase when the expected transition has not occurred. In this way, manipulation progress is represented explicitly as movement through a structured sequence of execution stages rather than as a collection of independent plans. We count replanning traces that explicitly name the current execution phase. The evolved share reaches approximately 14\% in base, 21\% in common-sense, 8\% in complex, and 7\% in visual tasks, compared with near-zero initial rates in those families.

\begin{figure}[ht]
    \centering
    \includegraphics[width=\linewidth]{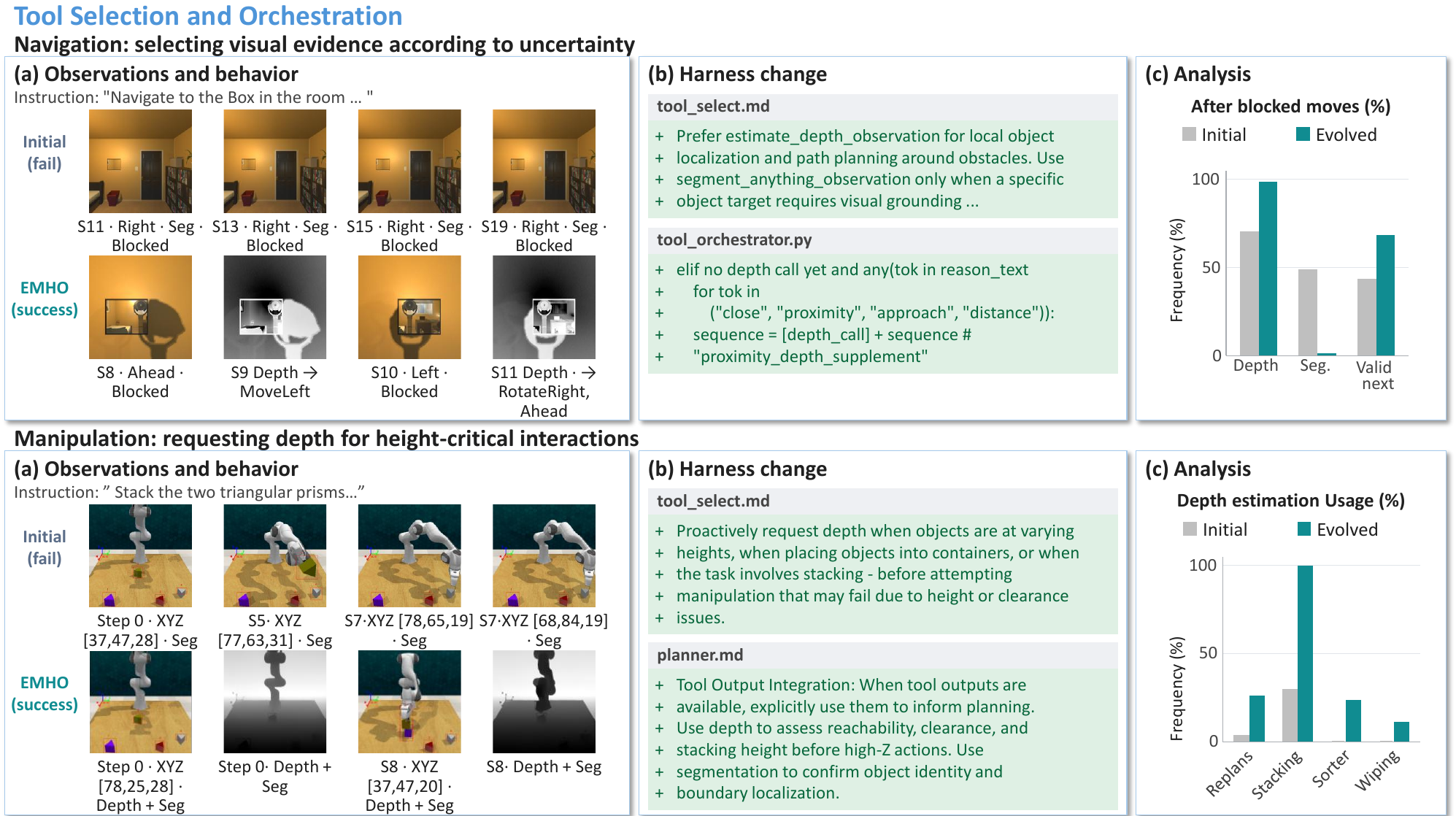}
    \caption{Qualitative analysis for tool usage in navigation and manipulation.}
    \label{fig:app-emho-qualitative-analysis-tool}
\end{figure}

\noindent\textbf{Tool Selection and Orchestration.} 
\label{sec:app_tool_orchestration} 

\noindent\emph{Navigation: selecting visual evidence according to uncertainty.} EMHO evolves the navigation harness to make tool use conditional on the information required for the next decision rather than invoking the same visual modules at every planning step. The resulting edits distinguish among appearance uncertainty, spatial or proximity uncertainty, and cases in which the current observation already provides sufficient evidence. The harness introduces explicit gating rules that favor no additional tool after a successful action when the scene and target direction remain clear, while retaining episode memory when recent blocked or failed actions make previous experience relevant. Tool-specific rules further route depth estimation toward obstacle and local path reasoning, segmentation toward identifying a particular target, and zoom toward resolving fine visual details once the relevant region is known. Complementary orchestration rules also invoke depth estimation when the planner expresses uncertainty about proximity or layout. We analyze tool calls immediately after blocked moves. Depth-estimation use rises from approximately 70\% to nearly 100\%, while segmentation use falls from approximately 50\% to near zero. The share of valid next actions rises from approximately 44\% to 68\%.

\noindent\emph{Manipulation: requesting depth for height-critical interactions.}
EMHO introduces an instruction to request depth proactively when object heights differ, the task involves stacking, or placement into a container requires clearance reasoning. In the illustrated stacking episode, the evolved harness requests depth together with segmentation both at the initial perception step and again during replanning, whereas the initial harness relies on segmentation alone. We further observe depth-estimation usage across task types and observe the largest increase for stacking tasks, where estimating the relative positions and heights of the object and the robot arm is particularly important for successful placement.

\begin{figure}[ht]
    \centering
    \includegraphics[width=\linewidth]{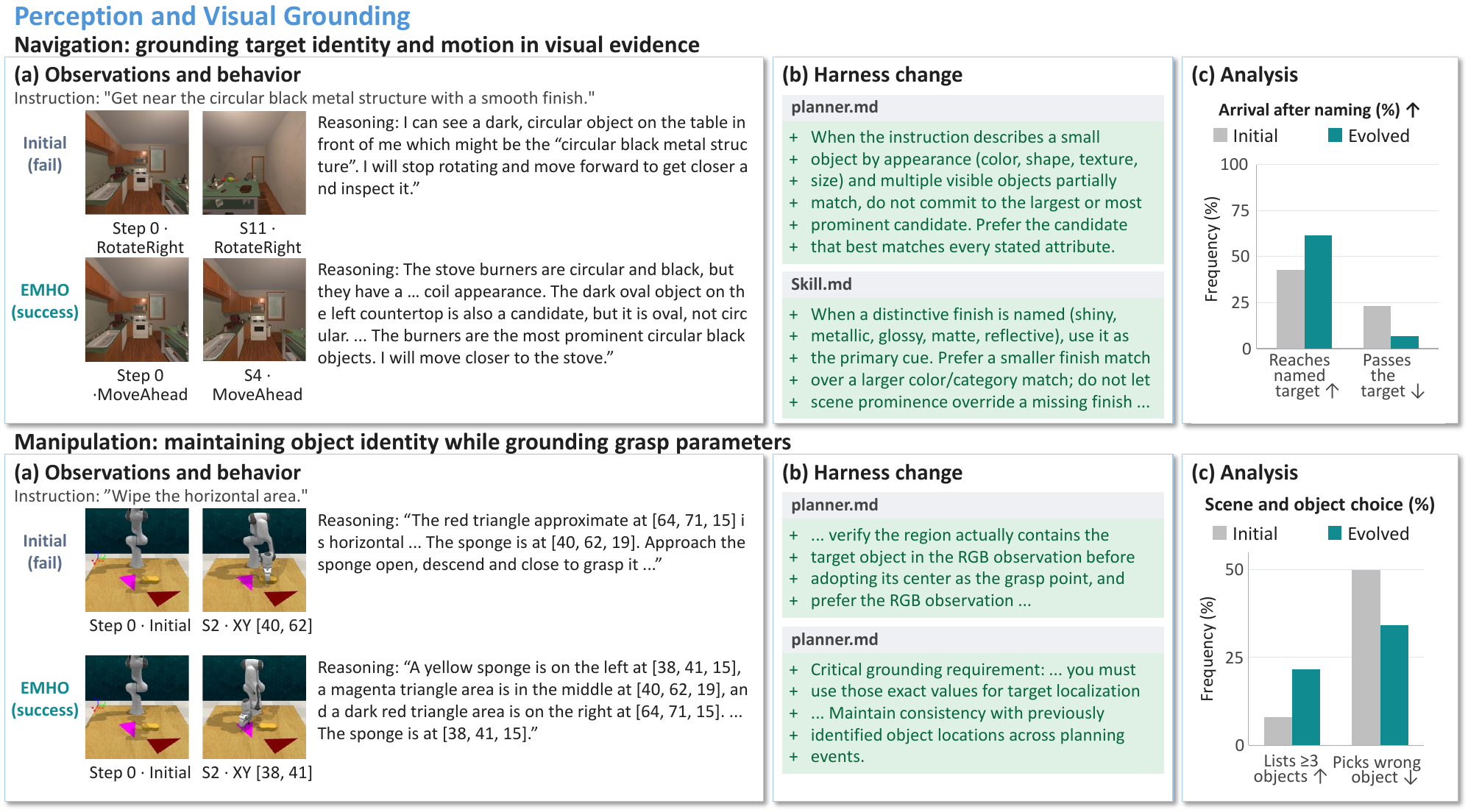}
    \caption{Qualitative analysis for perception and grounding in navigation and manipulation.}
    \label{fig:app-emho-qualitative-analysis-grounding}
\end{figure}
\noindent\textbf{Perception and Grounding.} 
\label{sec:app_visual_grounding} 

\noindent\emph{Navigation: grounding target identity and motion in visual evidence.} EMHO evolves the navigation harness to connect perceptual evidence more explicitly to both target selection and the subsequent motion decision. When several visible objects partially match an instruction, the planner is encouraged to compare the attributes specified by the task---such as color, material, shape, or finish---rather than choosing the most visually salient candidate. Other edits propagate the target phrase from the instruction directly into segmentation, allowing the visual query itself to reflect the object description, while the planner is instructed to treat segmentation, zoom, and the original RGB image as complementary sources of evidence. This also allows the harness to recover from imperfect tool outputs. When segmentation produces an empty or unreliable mask, the planner can retain a visually supported target hypothesis from the RGB observation rather than interpreting detector failure as evidence that the object is absent. The resulting trajectories exhibit clearer evidence-to-decision chains: visual tools help resolve target attributes, the planner compares candidate objects using those attributes, and the selected evidence is then carried forward into the navigation decision. 
To quantify this behavior, we measure the conditional arrival rate after the agent has explicitly identified a target in its reasoning. This rate increases from 43\% to 62\%. At the same time, the fraction of such trajectories in which the agent passes the target and subsequently moves farther away decreases from 23\% to 7\%. These results indicate improved navigation after a target hypothesis has been formed, while not establishing a general improvement in target-identification accuracy.

\noindent\emph{Manipulation: maintaining object identity while grounding grasp parameters.} EMHO evolves manipulation grounding around the distinction between identifying the intended object and choosing an executable grasp location. The harness requires relative descriptions such as ``the right object'' to be resolved against the current image, treats segmentation boxes and cropped regions as approximate perceptual cues rather than directly executable coordinates, and retains the RGB observation as the primary evidence for determining the final grasp pose. When replanning after an unsuccessful attempt, the agent can combine the current object list, newly requested visual evidence, and the history of previous grasp locations to re-estimate where the intended object is and modify the corresponding action parameters. In this way, grounding is treated as an ongoing process that can be updated from new observations rather than fixing the target location from a single initial detection. We inspect the planner's scene description and first object choice. Descriptions naming at least three objects with coordinates rise from approximately 8\% to 22\%, and first plans choosing the wrong object fall from 50\% to 34\% across evolved harnesses. These measurements support clearer scene representation and fewer relational object-choice errors, but do not establish fine-grained grasp-coordinate accuracy.

\subsection{Evolution Across Optimization Iterations}
\label{appendix:evolution_across_iterations}

\begin{figure}[t]
    \centering
    \begin{subfigure}[t]{0.47\linewidth}
        \centering
        \includegraphics[width=\linewidth]{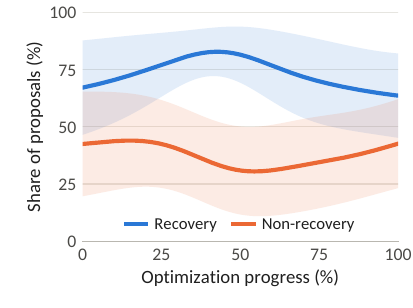}
        \caption{Navigation 27B}
        \label{fig:edit_flow_nav27}
    \end{subfigure}
    \hfill
    \begin{subfigure}[t]{0.47\linewidth}
        \centering
        \includegraphics[width=\linewidth]{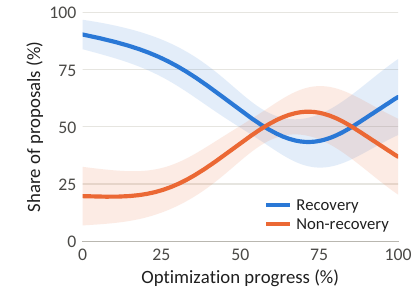}
        \caption{Manipulation 27B}
        \label{fig:edit_flow_manip27}
    \end{subfigure}

    \caption{
    Evolution of recovery and non-recovery edits across the harness optimization process. The relative composition of proposed edits changes
    differently across navigation and manipulation.
    }
    \label{fig:edit_evolution}
\end{figure}

\noindent\textbf{Recovery edits dominate early, while non-recovery edits emerge over time.} Figure~\ref{fig:edit_evolution} shows how the relative share of recovery and non-recovery edits changes over the course of optimization. In both domains, recovery edits are more prominent during the early iterations, suggesting that explicit or easily identifiable execution failures provide a natural initial target for harness refinement. However, the subsequent evolution differs across tasks. In navigation, recovery edits remain dominant throughout optimization; the share of non-recovery edits dips in the middle of optimization and recovers toward the end, narrowing the gap between the two categories. In manipulation, by contrast, the share of recovery edits decreases substantially as optimization proceeds, while non-recovery edits become increasingly prevalent during the middle and later iterations before the two categories converge.

\section{Detailed Results of Harness Modifications}
\label{sec:app_harness_modifications}

The examples below illustrate representative changes discovered for navigation and manipulation. For readability, some examples are shortened by omitting less relevant details.

\paragraph{Navigation.} 
Example of the evolved harness in the navigation task. 

%
%

\begingroup
\setlength{\fboxsep}{0pt}
\setlength{\parindent}{0pt}
\setlength{\parskip}{0pt}

\definecolor{NavInk}{HTML}{152A4A}
\definecolor{NavNumFG}{HTML}{405678}
\definecolor{NavAddFG}{HTML}{09683D}
\definecolor{NavAddBG}{HTML}{DFF0E4}
\definecolor{NavDelFG}{HTML}{BA2027}
\definecolor{NavDelBG}{HTML}{FCE1E3}
\definecolor{NavHeaderBG}{HTML}{DDE1E6}
\definecolor{NavHunkBG}{HTML}{EEF0F3}

\newlength{\NavSourceWidth}
\newlength{\NavInnerWidth}
\setlength{\NavSourceWidth}{610pt}
\setlength{\NavInnerWidth}{604pt}

\newcommand{\NavMono}{\ttfamily}
\newcommand{\NavMonoBold}{\ttfamily\bfseries}

\newcommand{\NavScaleRow}[1]{%
  \noindent\resizebox{\linewidth}{!}{#1}\par%
}

\newcommand{\NavFileHeader}[1]{%
  \NavScaleRow{%
    \colorbox{NavHeaderBG}{\makebox[\NavSourceWidth][l]{%
      \hspace*{8pt}\rule{0pt}{16.5pt}%
      \raisebox{0.8pt}{\NavMonoBold\fontsize{13pt}{13pt}\selectfont\color{NavInk}#1}%
    }}%
  }%
}

\newcommand{\NavDiffRow}[7]{%
  \NavScaleRow{%
    \colorbox{#1}{\makebox[\NavSourceWidth][l]{%
      \rule{0pt}{8.55pt}%
      \hspace*{6pt}%
      \makebox[8pt][c]{\NavMono\fontsize{11pt}{13pt}\selectfont\color{#2}#5}%
      \hspace*{1pt}\hspace*{#6pt}%
      \raisebox{0.1pt}{\NavMono\fontsize{10pt}{12pt}\selectfont\color{#2}#7}%
    }}%
  }%
}

\newcommand{\NavDiffRowWrap}[7]{%
  \NavScaleRow{%
    \colorbox{#1}{%
      \makebox[\NavSourceWidth][l]{%
        \rule{0pt}{8.55pt}%
        \hspace*{6pt}%
        \makebox[8pt][c]{%
          \NavMono\fontsize{11pt}{13pt}\selectfont\color{#2}#5%
        }%
        \hspace*{1pt}\hspace*{#6pt}%
        \parbox[t]{\dimexpr\NavSourceWidth-26pt-#6pt\relax}{%
          \vspace{0pt}%
          \setlength{\parindent}{0pt}%
          \setlength{\parskip}{0pt}%
          \raggedright
          \hyphenpenalty=10000%
          \exhyphenpenalty=10000%
          \NavMono\fontsize{10pt}{10.8pt}\selectfont\color{#2}#7\strut%
        }%
      }%
    }%
  }%
}

\newcommand{\NavHunkRow}[1]{%
  \NavScaleRow{%
    \colorbox{NavHunkBG}{\makebox[\NavSourceWidth][l]{%
      \hspace*{8pt}\rule{0pt}{9pt}%
      \NavMono\fontsize{7pt}{9pt}\selectfont\color{NavNumFG}#1%
    }}%
  }%
}

\newcommand{\NavGap}[1]{%
  \par\vspace*{#1pt}%
}

\newcommand{\NavEllipsis}[1]{%
  \NavScaleRow{%
    \makebox[\NavSourceWidth][c]{%
      \rule{0pt}{#1pt}%
      \raisebox{1pt}{\NavMonoBold\fontsize{11pt}{11pt}\selectfont\color{NavInk}...}%
    }%
  }%
}

\vspace{10pt}
\offinterlineskip
\NavFileHeader{harness/prompts/final\_planner.md}
\NavDiffRow{white}{NavInk}{2}{2}{}{0.0}{}
\NavDiffRow{white}{NavInk}{3}{3}{}{0.0}{Use the current visual observation, valid navigation actions, action history,}
\NavDiffRow{white}{NavInk}{4}{4}{}{0.0}{environment feedback, selected context, and any tool outputs to choose a}
\NavDiffRow{NavDelBG}{NavDelFG}{5}{}{-}{0.0}{short executable plan.}
\NavDiffRow{NavAddBG}{NavAddFG}{}{5}{+}{0.0}{short executable plan. Incorporate lessons from recent feedback, especially when}
\NavDiffRow{NavAddBG}{NavAddFG}{}{6}{+}{0.0}{earlier actions produced blocked or failed outcomes.}
\NavDiffRow{NavAddBG}{NavAddFG}{}{7}{+}{0.0}{}
\NavDiffRow{white}{NavInk}{6}{8}{}{0.0}{Do not choose an action from target visibility alone. If the target or relevant}
\NavDiffRow{white}{NavInk}{7}{9}{}{0.0}{object is visible, consider its apparent direction, distance, alignment,}
\NavDiffRow{white}{NavInk}{8}{10}{}{0.0}{obstacles, and recent feedback before deciding whether to move, rotate, tilt, or}
\NavDiffRow{NavDelBG}{NavDelFG}{9}{}{-}{0.0}{reposition.}
\NavDiffRow{NavAddBG}{NavAddFG}{}{11}{+}{0.0}{reposition.}
\NavDiffRow{NavAddBG}{NavAddFG}{}{12}{+}{0.0}{}
\NavDiffRowWrap{NavAddBG}{NavAddFG}{}{13}{+}{0.0}{When action history shows prior errors or blocked attempts with memory available, factor that into}
\NavDiffRowWrap{NavAddBG}{NavAddFG}{}{14}{+}{0.0}{your plan to avoid known failures. When action history shows prior successful steps, build on them}
\NavDiffRow{NavAddBG}{NavAddFG}{}{15}{+}{0.0}{rather than replanning unnecessarily.}
\NavDiffRow{NavAddBG}{NavAddFG}{}{16}{+}{0.0}{}
\NavDiffRow{NavAddBG}{NavAddFG}{}{17}{+}{0.0}{Minimize actions needed. Prefer shorter, direct paths unless intermediate actions are required}
\NavDiffRow{NavAddBG}{NavAddFG}{}{18}{+}{0.0}{to safely access the target.}
\NavDiffRow{NavAddBG}{NavAddFG}{}{19}{+}{0.0}{}
\NavDiffRow{NavAddBG}{NavAddFG}{}{20}{+}{0.0}{When planning new actions after successful steps, prioritize current observation over memory.}
\NavDiffRow{NavAddBG}{NavAddFG}{}{21}{+}{0.0}{}
\NavDiffRow{NavAddBG}{NavAddFG}{}{22}{+}{0.0}{Return only a JSON object with:}
\NavDiffRowWrap{NavAddBG}{NavAddFG}{}{23}{+}{0.0}{- "executable\_plan": a list of action objects, each with "action\_id" (integer) and "action\_name" (string)}
\NavDiffRow{NavAddBG}{NavAddFG}{}{24}{+}{0.0}{- Optional: "visual\_state\_description", "reasoning\_and\_reflection", "language\_plan"}
\NavDiffRow{NavAddBG}{NavAddFG}{}{25}{+}{0.0}{}
\NavDiffRow{NavAddBG}{NavAddFG}{}{26}{+}{0.0}{The planner will execute at most \{max\_plan\_actions\} actions.}
\NavGap{7}
\NavFileHeader{harness/planner.py}
\NavDiffRow{white}{NavInk}{4522}{4522}{}{26.04}{return [action\_id], json.dumps(reasoning)}
\NavDiffRow{white}{NavInk}{4523}{4523}{}{0.0}{}
\NavDiffRow{white}{NavInk}{4524}{4524}{}{19.22}{def \_invalid\_action\_response(self, error: str) -> Tuple[Any, str]:}
\NavDiffRow{NavAddBG}{NavAddFG}{}{4525}{+}{23.56}{if self.task\_type == "navigation":}
\NavDiffRow{NavAddBG}{NavAddFG}{}{4526}{+}{30.38}{valid\_actions = ", ".join(self.actions) if self.actions else "none"}
\NavDiffRow{NavAddBG}{NavAddFG}{}{4527}{+}{30.38}{reasoning = json.dumps(}
\NavDiffRow{NavAddBG}{NavAddFG}{}{4528}{+}{37.2}{\{}
\NavDiffRow{NavAddBG}{NavAddFG}{}{4529}{+}{44.02}{"visual\_state\_description": "Qwen Code returned an invalid action.",}
\NavDiffRow{NavAddBG}{NavAddFG}{}{4530}{+}{44.02}{"reasoning\_and\_reflection": (}
\NavDiffRow{NavAddBG}{NavAddFG}{}{4531}{+}{50.84}{f"\{error\} The previous action was not executed. "}
\NavDiffRow{NavAddBG}{NavAddFG}{}{4532}{+}{50.84}{f"Choose a different valid navigation action from: \{valid\_actions\}."}
\NavDiffRow{NavAddBG}{NavAddFG}{}{4533}{+}{44.02}{),}
\NavDiffRowWrap{NavAddBG}{NavAddFG}{}{4534}{+}{44.02}{"language\_plan": "No valid action was produced; select an alternative navigation action.",}
\NavDiffRow{NavAddBG}{NavAddFG}{}{4535}{+}{44.02}{"executable\_plan": [],}
\NavDiffRow{NavAddBG}{NavAddFG}{}{4536}{+}{37.2}{\}}
\NavDiffRow{NavAddBG}{NavAddFG}{}{4537}{+}{30.38}{)}
\NavDiffRow{NavAddBG}{NavAddFG}{}{4538}{+}{30.38}{return -1, reasoning}
\NavDiffRow{white}{NavInk}{4525}{4539}{}{26.04}{reasoning = json.dumps(}
\NavDiffRow{white}{NavInk}{4526}{4540}{}{32.86}{\{}
\NavDiffRow{white}{NavInk}{4527}{4541}{}{39.68}{"visual\_state\_description": "Qwen Code returned an invalid action.",}
\NavGap{11}
\NavFileHeader{harness/prompts/repair.md}
\NavDiffRow{white}{NavInk}{3}{3}{}{0.0}{The previous solver output did not match the required schema. Preserve the}
\NavDiffRow{white}{NavInk}{4}{4}{}{0.0}{current task context, valid action ids, and valid action names. Make the smallest}
\NavDiffRow{white}{NavInk}{5}{5}{}{0.0}{repair that restores the expected JSON shape.}
\NavDiffRow{NavAddBG}{NavAddFG}{}{6}{+}{0.0}{}
\NavDiffRow{NavAddBG}{NavAddFG}{}{7}{+}{0.0}{Common failure patterns to avoid:}
\NavDiffRow{NavAddBG}{NavAddFG}{}{8}{+}{0.0}{1. Wrapping output in markdown code blocks - unwrap to plain JSON}
\NavDiffRow{NavAddBG}{NavAddFG}{}{9}{+}{0.0}{2. Adding conversational text before/after JSON - output only the JSON object}
\NavDiffRow{NavAddBG}{NavAddFG}{}{10}{+}{0.0}{3. Missing executable\_plan key - ensure it exists as an array}
\NavDiffRow{NavAddBG}{NavAddFG}{}{11}{+}{0.0}{4. Invalid action\_id values - use only valid action ids from the available set}
\NavDiffRow{NavAddBG}{NavAddFG}{}{12}{+}{0.0}{5. Plan length exceeding max\_plan\_actions - output at most the configured maximum}
\NavDiffRowWrap{NavAddBG}{NavAddFG}{}{13}{+}{0.0}{6. Non-dict items in executable\_plan - each item must be an object with action\_id and action\_name}
\NavDiffRow{NavAddBG}{NavAddFG}{}{14}{+}{0.0}{7. Missing action\_id - every plan item must include an integer action\_id}
\NavDiffRow{NavAddBG}{NavAddFG}{}{15}{+}{0.0}{}
\NavDiffRow{NavAddBG}{NavAddFG}{}{16}{+}{0.0}{Output valid JSON only. Do not include markdown formatting or conversational text.}
\NavDiffRowWrap{NavAddBG}{NavAddFG}{}{17}{+}{0.0}{Ensure executable\_plan is a non-empty list of objects, each with valid action\_id and action\_name.}
\NavGap{11}
\NavFileHeader{harness/prompts/tool\_select.md}
\NavDiffRow{white}{NavInk}{1}{1}{}{0.0}{Selection guidance:}
\NavDiffRow{white}{NavInk}{2}{2}{}{0.0}{- Inspect the current observation image and the previous actions/environment feedback.}
\NavDiffRowWrap{white}{NavInk}{3}{3}{}{0.0}{- Decide which optional support would help the next task-solving step: episode memory or vision evidence.}
\NavDiffRow{NavDelBG}{NavDelFG}{4}{}{-}{0.0}{- Use memory when current-episode history or feedback is needed to avoid repeating mistakes.}
\NavDiffRowWrap{NavAddBG}{NavAddFG}{}{4}{+}{0.0}{- Use memory when current-episode history or feedback is needed to avoid repeating mistakes, especially when}
\NavDiffRow{NavAddBG}{NavAddFG}{}{5}{+}{0.0}{  recent actions produced blocked, failed, or unexpected feedback.}
\NavDiffRowWrap{white}{NavInk}{5}{6}{}{0.0}{- Select the available evidence source with the greatest expected marginal value for the next action.}
\NavDiffRowWrap{NavDelBG}{NavDelFG}{6}{}{-}{0.0}{- Consider no-tool as a valid choice. Prefer new, decision-relevant information over repeated or redundant}
\NavDiffRow{NavDelBG}{NavDelFG}{7}{}{-}{0.0}{  evidence.}
\NavDiffRowWrap{NavAddBG}{NavAddFG}{}{7}{+}{0.0}{- Consider no-tool as a valid choice. Prefer no-tool when current observation and feedback clearly indicate}
\NavDiffRow{NavAddBG}{NavAddFG}{}{8}{+}{0.0}{  the next action--this avoids unnecessary context overhead when confidence is high.}
\NavDiffRow{NavAddBG}{NavAddFG}{}{9}{+}{0.0}{- Prefer new, decision-relevant information over repeated or redundant evidence.}
\NavDiffRowWrap{NavAddBG}{NavAddFG}{}{10}{+}{0.0}{- Consult episode memory when past contexts suggest specific hazards or when action outcomes appeared ambiguous.}
\NavDiffRowWrap{NavAddBG}{NavAddFG}{}{11}{+}{0.0}{- Only select vision tools when they provide new spatial or structural information not visible in current}
\NavDiffRow{NavAddBG}{NavAddFG}{}{12}{+}{0.0}{  observation and not already represented in the context.}
\NavDiffRow{white}{NavInk}{8}{13}{}{0.0}{}
\NavDiffRow{white}{NavInk}{9}{14}{}{0.0}{Final planner will execute at most \{max\_plan\_actions\} actions.}
\NavDiffRow{NavAddBG}{NavAddFG}{}{15}{+}{0.0}{}
\NavDiffRowWrap{NavAddBG}{NavAddFG}{}{16}{+}{0.0}{When tool\_use == none (no context selected), the final planner has observe-only information and must}
\NavDiffRow{NavAddBG}{NavAddFG}{}{17}{+}{0.0}{complete navigation using observations alone.}
\NavGap{25}
\NavFileHeader{harness/SKILL.md}
\NavDiffRow{white}{NavInk}{8}{8}{}{0.0}{This is an embodied navigation task.}
\NavDiffRow{white}{NavInk}{9}{9}{}{0.0}{}
\NavDiffRow{white}{NavInk}{10}{10}{}{0.0}{Use only the active task instruction, current observation images, valid action}
\NavDiffRow{NavDelBG}{NavDelFG}{11}{}{-}{0.0}{list, visible action history and feedback, context materialized by the harness,}
\NavDiffRow{NavAddBG}{NavAddFG}{}{11}{+}{0.0}{space, visible action history and feedback, context materialized by the harness,}
\NavDiffRow{white}{NavInk}{12}{12}{}{0.0}{and permitted vision-tool results. Never use hidden simulator state, benchmark}
\NavDiffRow{NavDelBG}{NavDelFG}{13}{}{-}{0.0}{source code, shortest paths, target coordinates, reward labels, or final success}
\NavDiffRow{NavDelBG}{NavDelFG}{14}{}{-}{0.0}{labels.}
\NavDiffRow{NavAddBG}{NavAddFG}{}{13}{+}{0.0}{source code, maps, shortest paths, target coordinates, object poses, reward}
\NavDiffRow{NavAddBG}{NavAddFG}{}{14}{+}{0.0}{labels, or final success labels.}
\NavDiffRow{white}{NavInk}{15}{15}{}{0.0}{}
\NavDiffRow{white}{NavInk}{16}{16}{}{0.0}{For a selector request, return only the requested module-selection JSON. Include}
\NavDiffRow{white}{NavInk}{17}{17}{}{0.0}{vision tool requests only in the selector field required by its schema.}
\endgroup

\paragraph{Manipulation.}
Example of the evolved harness in the manipulation task.

%
%

\begingroup
\setlength{\fboxsep}{0pt}
\setlength{\parindent}{0pt}
\setlength{\parskip}{0pt}

\definecolor{ManipInk}{HTML}{152A4A}
\definecolor{ManipNumFG}{HTML}{405678}
\definecolor{ManipAddFG}{HTML}{09683D}
\definecolor{ManipAddBG}{HTML}{DFF0E4}
\definecolor{ManipDelFG}{HTML}{BA2027}
\definecolor{ManipDelBG}{HTML}{FCE1E3}
\definecolor{ManipHeaderBG}{HTML}{DDE1E6}
\definecolor{ManipHunkBG}{HTML}{EEF0F3}

\newlength{\ManipSourceWidth}
\newlength{\ManipInnerWidth}
\setlength{\ManipSourceWidth}{612.375pt}
\setlength{\ManipInnerWidth}{606.375pt}

\newcommand{\ManipMono}{\ttfamily}
\newcommand{\ManipMonoBold}{\ttfamily\bfseries}

\newcommand{\ManipScaleRow}[1]{%
  \noindent\resizebox{\linewidth}{!}{#1}\par%
}

\newcommand{\ManipFileHeader}[1]{%
  \ManipScaleRow{%
    \colorbox{ManipHeaderBG}{\makebox[\ManipSourceWidth][l]{%
      \hspace*{8pt}\rule{0pt}{16.5pt}%
      \raisebox{0.8pt}{\ManipMonoBold\fontsize{13pt}{13pt}\selectfont\color{ManipInk}#1}%
    }}%
  }%
}

\newcommand{\ManipDiffRow}[7]{%
  \ManipScaleRow{%
    \colorbox{#1}{\makebox[\ManipSourceWidth][l]{%
      \rule{0pt}{8.55pt}%
      \hspace*{6pt}%
      \makebox[8pt][c]{\ManipMono\fontsize{11pt}{13pt}\selectfont\color{#2}#5}%
      \hspace*{1pt}\hspace*{#6pt}%
      \raisebox{0.1pt}{\ManipMono\fontsize{10pt}{12pt}\selectfont\color{#2}#7}%
    }}%
  }%
}

\newcommand{\ManipHunkRow}[1]{%
  \ManipScaleRow{%
    \colorbox{ManipHunkBG}{\makebox[\ManipSourceWidth][l]{%
      \hspace*{8pt}\rule{0pt}{9pt}%
      \ManipMono\fontsize{7pt}{9pt}\selectfont\color{ManipNumFG}#1%
    }}%
  }%
}

\newcommand{\ManipGap}[1]{%
  \par\vspace*{#1pt}%
}

\newcommand{\ManipEllipsis}[1]{%
  \ManipScaleRow{%
    \makebox[\ManipSourceWidth][c]{%
      \rule{0pt}{#1pt}%
      \raisebox{1pt}{\ManipMonoBold\fontsize{11pt}{11pt}\selectfont\color{ManipInk}...}%
    }%
  }%
}

\vspace{10pt}
\offinterlineskip
\ManipFileHeader{harness/prompts/final\_planner.md}
\ManipDiffRow{white}{ManipInk}{16}{16}{}{0.0}{`"[[X, Y, Z, Roll, Pitch, Yaw, Gripper]]"`.}
\ManipDiffRow{white}{ManipInk}{17}{17}{}{0.0}{Use the original EmbodiedBench manipulation rhythm when useful: approach open,}
\ManipDiffRow{white}{ManipInk}{18}{18}{}{0.0}{descend and close, lift closed, move/place, release open.}
\ManipDiffRow{ManipAddBG}{ManipAddFG}{}{19}{+}{0.0}{When re-planning after a previous plan exhausted without success, do not simply}
\ManipDiffRow{ManipAddBG}{ManipAddFG}{}{20}{+}{0.0}{re-emit the same approach/descent/lift sequence. First compare the current}
\ManipDiffRow{ManipAddBG}{ManipAddFG}{}{21}{+}{0.0}{visual state to the previous plan's intent: if the gripper is already above or}
\ManipDiffRow{ManipAddBG}{ManipAddFG}{}{22}{+}{0.0}{near the target, or the target is already grasped, start the new plan from the}
\ManipDiffRow{ManipAddBG}{ManipAddFG}{}{23}{+}{0.0}{next phase (e.g., lift, translate, release) instead of repeating the phase}
\ManipDiffRow{ManipAddBG}{ManipAddFG}{}{24}{+}{0.0}{that already completed.}
\ManipDiffRow{white}{ManipInk}{19}{25}{}{0.0}{Every X/Y/Z value must be in [0, 100], every rotation value must be in [0, 120],}
\ManipDiffRow{white}{ManipInk}{20}{26}{}{0.0}{and Gripper must be exactly 0 or 1.}
\ManipGap{9}
\ManipFileHeader{harness/planner.py}
\ManipDiffRow{white}{ManipInk}{4909}{4920}{}{31.0}{return True}
\ManipDiffRow{white}{ManipInk}{4910}{4921}{}{24.18}{if len(recent\_actions) >= 4 and recent\_actions[-4:-2] == recent\_actions[-2:]:}
\ManipDiffRow{white}{ManipInk}{4911}{4922}{}{31.0}{return True}
\ManipDiffRow{ManipAddBG}{ManipAddFG}{}{4923}{+}{24.18}{\# Near-identical replan loop: the planner re-issues the same action}
\ManipDiffRow{ManipAddBG}{ManipAddFG}{}{4924}{+}{24.18}{\# structure with only minor coordinate tweaks (e.g. grasp x 48 -> 41)}
\ManipDiffRow{ManipAddBG}{ManipAddFG}{}{4925}{+}{24.18}{\# after a failed attempt, so exact-repeat detection above misses it.}
\ManipDiffRow{ManipAddBG}{ManipAddFG}{}{4926}{+}{24.18}{recent\_vectors = [}
\ManipDiffRow{ManipAddBG}{ManipAddFG}{}{4927}{+}{31.0}{self.\_numeric\_action\_vector(record.get("action"))}
\ManipDiffRow{ManipAddBG}{ManipAddFG}{}{4928}{+}{31.0}{for record in self.\_executed\_action\_records[-6:]}
\ManipDiffRow{ManipAddBG}{ManipAddFG}{}{4929}{+}{24.18}{]}
\ManipDiffRow{ManipAddBG}{ManipAddFG}{}{4930}{+}{24.18}{recent\_vectors = [vector for vector in recent\_vectors if vector is not None]}
\ManipDiffRow{ManipAddBG}{ManipAddFG}{}{4931}{+}{24.18}{if len(recent\_vectors) >= 3:}
\ManipDiffRow{ManipAddBG}{ManipAddFG}{}{4932}{+}{31.0}{first, second, third = (}
\ManipDiffRow{ManipAddBG}{ManipAddFG}{}{4933}{+}{37.82}{recent\_vectors[-3],}
\ManipDiffRow{ManipAddBG}{ManipAddFG}{}{4934}{+}{37.82}{recent\_vectors[-2],}
\ManipDiffRow{ManipAddBG}{ManipAddFG}{}{4935}{+}{37.82}{recent\_vectors[-1],}
\ManipDiffRow{ManipAddBG}{ManipAddFG}{}{4936}{+}{31.0}{)}
\ManipDiffRow{ManipAddBG}{ManipAddFG}{}{4937}{+}{31.0}{if len(first) == len(second) == len(third):}
\ManipDiffRow{ManipAddBG}{ManipAddFG}{}{4938}{+}{37.82}{deltas = [}
\ManipDiffRow{ManipAddBG}{ManipAddFG}{}{4939}{+}{44.64}{max(}
\ManipDiffRow{ManipAddBG}{ManipAddFG}{}{4940}{+}{51.46}{abs(a - b)}
\ManipDiffRow{ManipAddBG}{ManipAddFG}{}{4941}{+}{51.46}{for a, b in zip(first, second)}
\ManipDiffRow{ManipAddBG}{ManipAddFG}{}{4942}{+}{44.64}{),}
\ManipDiffRow{ManipAddBG}{ManipAddFG}{}{4943}{+}{44.64}{max(}
\ManipDiffRow{ManipAddBG}{ManipAddFG}{}{4944}{+}{51.46}{abs(a - b)}
\ManipDiffRow{ManipAddBG}{ManipAddFG}{}{4945}{+}{51.46}{for a, b in zip(second, third)}
\ManipDiffRow{ManipAddBG}{ManipAddFG}{}{4946}{+}{44.64}{),}
\ManipDiffRow{ManipAddBG}{ManipAddFG}{}{4947}{+}{44.64}{max(}
\ManipDiffRow{ManipAddBG}{ManipAddFG}{}{4948}{+}{51.46}{abs(a - b)}
\ManipDiffRow{ManipAddBG}{ManipAddFG}{}{4949}{+}{51.46}{for a, b in zip(first, third)}
\ManipDiffRow{ManipAddBG}{ManipAddFG}{}{4950}{+}{44.64}{),}
\ManipDiffRow{ManipAddBG}{ManipAddFG}{}{4951}{+}{37.82}{]}
\ManipDiffRow{ManipAddBG}{ManipAddFG}{}{4952}{+}{37.82}{if all(delta <= 10.0 for delta in deltas):}
\ManipDiffRow{ManipAddBG}{ManipAddFG}{}{4953}{+}{44.64}{return True}
\ManipDiffRow{white}{ManipInk}{4912}{4954}{}{24.18}{return False}
\ManipDiffRow{white}{ManipInk}{4913}{4955}{}{0.0}{}
\ManipEllipsis{25}
\ManipDiffRow{white}{ManipInk}{4899}{4899}{}{31.0}{f"\{action\_text\}"}
\ManipDiffRow{white}{ManipInk}{4900}{4900}{}{24.18}{)}
\ManipDiffRow{white}{ManipInk}{4901}{4901}{}{0.0}{}
\ManipDiffRow{ManipAddBG}{ManipAddFG}{}{4902}{+}{17.36}{@staticmethod}
\ManipDiffRow{ManipAddBG}{ManipAddFG}{}{4903}{+}{17.36}{def \_numeric\_action\_vector(action: Any) -> Optional[List[float]]:}
\ManipDiffRow{ManipAddBG}{ManipAddFG}{}{4904}{+}{24.18}{if not isinstance(action, (list, tuple)) or len(action) < 3:}
\ManipDiffRow{ManipAddBG}{ManipAddFG}{}{4905}{+}{31.0}{return None}
\ManipDiffRow{ManipAddBG}{ManipAddFG}{}{4906}{+}{24.18}{values: List[float] = []}
\ManipDiffRow{ManipAddBG}{ManipAddFG}{}{4907}{+}{24.18}{for value in action:}
\ManipDiffRow{ManipAddBG}{ManipAddFG}{}{4908}{+}{31.0}{if isinstance(value, bool) or not isinstance(value, (int, float)):}
\ManipDiffRow{ManipAddBG}{ManipAddFG}{}{4909}{+}{37.82}{return None}
\ManipDiffRow{ManipAddBG}{ManipAddFG}{}{4910}{+}{31.0}{values.append(float(value))}
\ManipDiffRow{ManipAddBG}{ManipAddFG}{}{4911}{+}{24.18}{return values}
\ManipDiffRow{ManipAddBG}{ManipAddFG}{}{4912}{+}{0.0}{}
\ManipDiffRow{white}{ManipInk}{4902}{4913}{}{17.36}{def \_recent\_repeated\_loop(self) -> bool:}
\ManipDiffRow{white}{ManipInk}{4903}{4914}{}{24.18}{recent\_actions = [}
\ManipDiffRow{white}{ManipInk}{4904}{4915}{}{31.0}{self.\_action\_repeat\_key(record.get("action\_id"))}
\ManipGap{9}
\ManipFileHeader{harness/tool\_orchestrator.py}
\ManipDiffRow{white}{ManipInk}{129}{129}{}{33.48}{or "Use complementary visual evidence before planning."}
\ManipDiffRow{white}{ManipInk}{130}{130}{}{26.66}{)}
\ManipDiffRow{white}{ManipInk}{131}{131}{}{26.66}{override\_marker = self.\_path\_failure\_marker(observed\_evidence)}
\ManipDiffRow{ManipAddBG}{ManipAddFG}{}{132}{+}{0.0}{}
\ManipDiffRow{ManipAddBG}{ManipAddFG}{}{133}{+}{29.76}{\# Redundancy guard: if the same vision tool already ran successfully}
\ManipDiffRow{ManipAddBG}{ManipAddFG}{}{134}{+}{29.76}{\# in a recent planning event of this episode and no fresh}
\ManipDiffRow{ManipAddBG}{ManipAddFG}{}{135}{+}{29.76}{\# environment feedback (e.g. a path-planning failure) has arrived}
\ManipDiffRow{ManipAddBG}{ManipAddFG}{}{136}{+}{29.76}{\# since, skip re-issuing it. The planner keeps its RGB frame and}
\ManipDiffRow{ManipAddBG}{ManipAddFG}{}{137}{+}{29.76}{\# recent feedback, so a repeated identical segmentation call adds}
\ManipDiffRow{ManipAddBG}{ManipAddFG}{}{138}{+}{29.76}{\# no new evidence while consuming the per-plan vision budget.}
\ManipDiffRow{ManipAddBG}{ManipAddFG}{}{139}{+}{29.76}{if not override\_marker:}
\ManipDiffRow{ManipAddBG}{ManipAddFG}{}{140}{+}{36.58}{redundant\_tool = self.\_redundant\_vision\_tool(}
\ManipDiffRow{ManipAddBG}{ManipAddFG}{}{141}{+}{43.4}{requested\_tools,}
\ManipDiffRow{ManipAddBG}{ManipAddFG}{}{142}{+}{43.4}{episode\_state,}
\ManipDiffRow{ManipAddBG}{ManipAddFG}{}{143}{+}{43.4}{history,}
\ManipDiffRow{ManipAddBG}{ManipAddFG}{}{144}{+}{36.58}{)}
\ManipDiffRow{ManipAddBG}{ManipAddFG}{}{145}{+}{36.58}{if redundant\_tool:}
\ManipDiffRow{ManipAddBG}{ManipAddFG}{}{146}{+}{43.4}{return \{}
\ManipDiffRow{ManipAddBG}{ManipAddFG}{}{147}{+}{50.22}{"vision\_orchestration\_diagnostics": \{}
\ManipDiffRow{ManipAddBG}{ManipAddFG}{}{148}{+}{57.04}{"source": "expanded\_manipulation\_harness",}
\ManipDiffRow{ManipAddBG}{ManipAddFG}{}{149}{+}{57.04}{"tool": "",}
\ManipDiffRow{ManipAddBG}{ManipAddFG}{}{150}{+}{57.04}{"requested\_tool": requested\_tool,}
\ManipDiffRow{ManipAddBG}{ManipAddFG}{}{151}{+}{57.04}{"requested\_tools": requested\_tools,}
\ManipDiffRow{ManipAddBG}{ManipAddFG}{}{152}{+}{57.04}{"final\_tools": [],}
\ManipDiffRow{ManipAddBG}{ManipAddFG}{}{153}{+}{57.04}{"sequence\_length": 0,}
\ManipDiffRow{ManipAddBG}{ManipAddFG}{}{154}{+}{57.04}{"max\_calls": self.max\_calls,}
\ManipDiffRow{ManipAddBG}{ManipAddFG}{}{155}{+}{57.04}{"routing\_source": "redundancy\_guard",}
\ManipDiffRow{ManipAddBG}{ManipAddFG}{}{156}{+}{57.04}{"selector\_overridden": False,}
\ManipDiffRow{ManipAddBG}{ManipAddFG}{}{157}{+}{57.04}{"override\_marker": "",}
\ManipDiffRow{ManipAddBG}{ManipAddFG}{}{158}{+}{57.04}{"selected\_by\_selector": selected,}
\ManipDiffRow{ManipAddBG}{ManipAddFG}{}{159}{+}{57.04}{"reason": (}
\ManipDiffRow{ManipAddBG}{ManipAddFG}{}{160}{+}{63.86}{f"Skipped \{redundant\_tool\}: the same tool "}
\ManipDiffRow{ManipAddBG}{ManipAddFG}{}{161}{+}{63.86}{"already delivered a successful result in a "}
\ManipDiffRow{ManipAddBG}{ManipAddFG}{}{162}{+}{63.86}{"recent planning event and no fresh environment "}
\ManipDiffRow{ManipAddBG}{ManipAddFG}{}{163}{+}{63.86}{"feedback has arrived since."}
\ManipDiffRow{ManipAddBG}{ManipAddFG}{}{164}{+}{57.04}{)[:500],}
\ManipDiffRow{ManipAddBG}{ManipAddFG}{}{165}{+}{57.04}{"selector\_reason": selector\_reason[:500],}
\ManipDiffRow{ManipAddBG}{ManipAddFG}{}{166}{+}{50.22}{\}}
\ManipDiffRow{ManipAddBG}{ManipAddFG}{}{167}{+}{43.4}{\}}
\ManipEllipsis{23}
\ManipDiffRow{ManipAddBG}{ManipAddFG}{}{168}{+}{0.0}{}
\ManipDiffRow{ManipAddBG}{ManipAddFG}{}{369}{+}{52.08}{\# Check that no fresh feedback arrived after this call.}
\ManipDiffRow{ManipAddBG}{ManipAddFG}{}{370}{+}{52.08}{\# The entry's own env\_feedback is the feedback that was}
\ManipDiffRow{ManipAddBG}{ManipAddFG}{}{371}{+}{52.08}{\# present when the tool ran; if the latest history entry}
\ManipDiffRow{ManipAddBG}{ManipAddFG}{}{372}{+}{52.08}{\# has different (non-empty) feedback, the scene may have}
\ManipDiffRow{ManipAddBG}{ManipAddFG}{}{373}{+}{52.08}{\# changed and the call is not redundant.}
\ManipDiffRow{ManipAddBG}{ManipAddFG}{}{374}{+}{52.08}{entry\_feedback = str(entry.get("env\_feedback", "")).strip()}
\ManipDiffRow{ManipAddBG}{ManipAddFG}{}{375}{+}{52.08}{latest = history[-1] if history else \{\}}
\ManipDiffRow{ManipAddBG}{ManipAddFG}{}{376}{+}{52.08}{latest\_feedback = (}
\ManipDiffRow{ManipAddBG}{ManipAddFG}{}{377}{+}{58.9}{str(latest.get("env\_feedback", "")).strip()}
\ManipDiffRow{ManipAddBG}{ManipAddFG}{}{378}{+}{58.9}{if isinstance(latest, dict)}
\ManipDiffRow{ManipAddBG}{ManipAddFG}{}{379}{+}{58.9}{else ""}
\ManipDiffRow{ManipAddBG}{ManipAddFG}{}{380}{+}{52.08}{)}
\ManipDiffRow{ManipAddBG}{ManipAddFG}{}{381}{+}{52.08}{if latest\_feedback and latest\_feedback != entry\_feedback:}
\ManipDiffRow{ManipAddBG}{ManipAddFG}{}{382}{+}{58.9}{continue}
\ManipDiffRow{ManipAddBG}{ManipAddFG}{}{383}{+}{52.08}{return tool\_name}
\ManipDiffRow{ManipAddBG}{ManipAddFG}{}{384}{+}{29.76}{return ""}
\endgroup

\end{document}